\documentclass[sigconf,anonymous=False]{acmart}
\usepackage{booktabs}
\usepackage{multirow}  
\usepackage{subfigure}

\usepackage{rotating}
\usepackage{graphicx}
\usepackage{lipsum}
\usepackage{textcomp}

\setcopyright{acmcopyright}
\copyrightyear{2020}
\acmYear{2020}
\acmDOI{10.1145/1122445.1122456}

\acmConference[IoTDI '21]{IoTDI '21: The 6th ACM/IEEE Conference on Internet of Things Design and Implementation}{May 18--21, 2021}{Nashville, USA}

\acmPrice{15.00}
\acmISBN{978-1-4503-7005-9/20/11}
\begin{document}

\title{Transfer Learning for Thermal Comfort Prediction in Multiple Cities}

\author{Nan Gao}
\email{nan.gao@rmit.edu.au}
\affiliation{%
  \institution{RMIT University}
  \city{Melbourne}
  \country{Australia}
  \postcode{3000}
}
\author{Wei Shao}
\email{wei.shao@rmit.edu.au}
\affiliation{%
  \institution{RMIT University}
  \city{Melbourne}
  \country{Australia}
  \postcode{3000}
}
\author{Mohammad Saiedur Rahaman}
\email{saiedur.rahaman@rmit.edu.au}
\affiliation{%
  \institution{RMIT University}
  \city{Melbourne}
  \country{Australia}
  \postcode{3000}
}
\author{Jun Zhai}
\email{jun.zhai@rmit.edu.au}
\affiliation{%
  \institution{RMIT University}
  \city{Melbourne}
  \country{Australia}
  \postcode{3000}
}

\author{Klaus David}
\email{david@uni-kassel.de}
\affiliation{%
  \institution{University of Kassel}
  \city{Kassel}
  \country{Germany}
}

\author{Flora D. Salim}
\email{flora.salim@rmit.edu.au}
\affiliation{%
  \institution{RMIT University}
  \city{Melbourne}
  \country{Australia}
  \postcode{3000}
}

\renewcommand{\shortauthors}{Nan and Wei, et al.}

\begin{abstract}
HVAC (Heating, Ventilation and Air Conditioning) system is an important part of a building, which constitutes up to 40\% of building energy usage. The main purpose of HVAC, maintaining appropriate thermal comfort, is crucial for the best utilisation of energy usage. Besides, thermal comfort is also crucial for well-being, health, and work productivity. Recently, data-driven thermal comfort models have got better performance than traditional knowledge-based methods (e.g. Predicted Mean Vote Model). An accurate thermal comfort model requires a large amount of self-reported thermal comfort data from indoor occupants which undoubtedly remains a challenge for researchers. In this research, we aim to tackle this data-shortage problem and boost the performance of thermal comfort prediction. We utilise sensor data from multiple cities in the same climate zone to learn thermal comfort patterns. We present a transfer learning based multilayer perceptron model from the same climate zone (TL-MLP-C*) for accurate thermal comfort prediction. Extensive experimental results on ASHRAE RP-884, the Scales Project and Medium US Office datasets show that the performance of the proposed TL-MLP-C* exceeds the state-of-the-art methods in accuracy and F1-score. 
\end{abstract}



\maketitle

\section{Introduction}
\label{sec:intro}
Internet of Things (IoT) devices have been widely used in the urban environment across various disciplines such as early-warning systems, traffic management, environment monitoring and buildings' HVAC (Heating, Ventilation and Air Conditioning) systems \cite{hu2015smartroad,abbas2018inverted,elnour2020sensor}. At the same time, sensors have become the backbones of smart cities which enable spatial and situational awareness of real-time monitoring in dynamic phenomena, e.g., pedestrian movement \cite{wang2017predicting}, parking events \cite{shao2018parking,shao2017traveling}, occupancy recognition \cite{ang2016human} and energy consumption \cite{song2019evolutionary}. 

As one of the most important parts in cities all over the world, buildings account for about 40\% global energy usage and 60\% worldwide electricity usage \cite{sadid2017discrete}. A large proportion of this usage is contributed by the buildings' HVAC system. In sub-tropical climate cities like Sydney, HVAC even consumes about 70\% of the buildings' energy usage \cite{rahman2010energy,zhang2018thermal}. With different kinds of IoT sensors (e.g., temperature sensor, humidity sensor, air velocity sensor, air quality sensor) installed in the building, the HVAC system can dynamically maintain the indoor occupant comfort at minimal energy usage. To achieve overall satisfaction with the indoor environment, thermal comfort is often considered to be the most influential factor compared with other factors such as visual and acoustics comfort \cite{frontczak2011literature}.

\textit{Thermal Comfort} is the state of mind which expresses satisfaction with the thermal environment (ASHRAE Standard 2004 \cite{american2004thermal}). Researchers have found that thermal discomfort not only affects occupant productivity, work performance and engagement \cite{gao2020n}, but also has a bad influence on lifelong health. Hence, it is important to maintain a thermal-comfort environment for the well-being of the occupants while minimizing the buildings' energy usage. A crucial step towards this goal is to create an accurate model for thermal comfort.

\begin{figure}[t]
    \centering
    \includegraphics[width=0.38\textwidth]{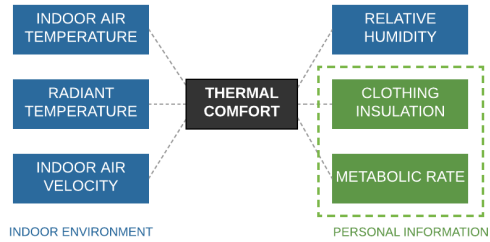}
    \caption{Six factors affecting thermal comfort (PMV model)}
    \label{fig:pmv}
\end{figure}

The \textit{Predicted Mean Vote} (PMV) model proposed by Fanger et al. \cite{fanger1970thermal} stands among the most prevalent thermal comfort model, which was developed with principles of human heat-balance and adopted by the American Society of Heating, Refrigerating and Air-conditioning Engineers (ASHRAE) Standard 55. The PMV model relates thermal comfort scale with six different factors: air temperature, mean radiant temperature, relative humidity, air speed, metabolic rate, and clothing insulation (see Figure \ref{fig:pmv}). Then, the average thermal sensation score can be calculated at a 7-point scale ranging from 3 to -3, which indicates feeling hot, warm, slightly warm, neutral, slightly cool, cool and cold. 

However, some researchers revealed the discrepancy between thermal sensation vote reported by occupants and predicted mean vote \cite{becker2009thermal,Hoof2008FortyYO}. It could probably be that a variety of parameters may affect thermal comfort such as time factors (e.g., hour, day, season) \cite{auffenberg2015personalised, chun2008thermal}, personal information  (e.g., heart rate, skin temperature, age, gender)  \cite{chaudhuri2018random, indraganti2015thermal,indraganti2010effect,karjalainen2007gender}, environmental factors (e.g., air quality, color, light, noise, outdoor climates)  \cite{seppanen1999association,kolokotsa2001advanced}, culture (e.g., dress code, economic status) \cite{indraganti2010effect}, short and long-time thermal exposure \cite{chun2008thermal}, etc. Therefore, the data-driven method is a better choice compared with traditional PMV model as more parameters could be utilized to improve the performance of thermal comfort prediction. 

To build a data-driven thermal comfort model, there are mainly two types of approaches for obtaining occupants' thermal comfort feelings in the buildings. One is the survey-based approach \cite{di2020era5, nikolaou2009virtual, langevin2015tracking}, which utilizes a participatory learning process with a questionnaire. The other one is the physiological measurement-based approach \cite{liu2018personal,choi2019development}. It records certain physiological signals (skin temperature \cite{liu2011evaluation}, heart rate \cite{liu2008heart}, skin conductance \cite{gerrett2013comparison}, etc.) from wearable sensors. However, both approaches need real-time and continuous monitoring or feedback of indoor occupants, which can be regarded as a burden for participants, making it a challenging task for researchers, especially for those with limited time and budget. Considering the definition of thermal comfort is 'the state of mind which expresses satisfaction of thermal environment', the survey-based approaches potentially learn personal thermal comfort more accurately than physiological approaches as they try to directly extract the state of mind of a person. Thus, in this research, we will explore ways to deal with the lack of thermal comfort survey data.

Various thermal comfort studies have been carried out in different cities all over the world, and several databases including multiple cities and climate zones are currently online (see Section ~\ref{sec: dataset introduction}). Considered the sensor data inferred from different cities may have very divergent patterns caused by the building materials, construction requirements and climate changes, previous studies mainly focus on investigating how people group living in specific cities react to their thermal environment such as hot-arid climate in Kalgoorlie-Boulder Australia  \cite{cena1999field}, humid subtropical climate in Brisbane Australia \cite{de1998developing}.

Recently, there are many publications applying machine learning algorithms to predict thermal comfort for a specified group of people (e.g., like for a group working in the same building). However, it is usually hard to get enough labelled thermal comfort data, which limits the performance of data-driven thermal comfort modelling. The considered occupant thermal comfort level has a strong correlation with indoor environment sensor data (e.g., air temperature, air velocity, humidity and radiant temperature) and physiological data (e.g., metabolic rate, skin temperature). It is possible to utilize sensor data from multiple cities to benefit the target building in another city. 

Therefore, in this paper, we hypothesize that the performance of thermal comfort prediction can be boosted by transfer learning across sensor data from multiple cities. Then, we aim to answer the following research questions: \textit{Can we predict occupants' thermal comfort accurately by learning from multiple buildings in the same climate zone when we do not have enough data? If so, which features contribute most for effective thermal comfort transfer learning? } 

To answer the above questions, we present a thermal comfort prediction framework with transfer learning technique, which aims to predict occupants' thermal sensation with insufficient labelled data. \textit{ASHRAE RP-884} database \cite{fanger1970thermal} and \textit{the Scales Project} dataset \cite{schweiker2019scales} are chosen as source datasets and \textit{Medium US Office} \cite{langevin2015tracking} is used as the target dataset. To deal with the imbalanced class problem, we merge the minority classes and divide the data into five categories, then apply a generative adversarial network based resampling method \textit{TabularGAN} for meaningful thermal comfort classification. The thermal comfort prediction model for the target dataset is trained by retaining the last hidden layer of the neural network from the source domain. 
To summarize, we make the following contributions as listed:

\begin{itemize}
    \item To the best of our knowledge, we are the first to transfer the knowledge from similar thermal environments (climate zones) to the target building for effective thermal comfort modelling. We propose the transfer learning based multilayer perceptron (TL-MLP) model and transfer learning based multilayer perceptron from the same climate zone (TL-MLP-C*) model. We confirm that thermal comfort sensor data from multiple cities in the same climate zone can benefit the small thermal comfort dataset for the target building in another city with insufficient training data. 
    \item Extensive experimental results show that the proposed TL-MLP and TL-MLP-C* models outperform the state-of-the-art algorithms for thermal comfort prediction and can be implemented in any building without adequate thermal comfort labelled data. 
    \item We identify the most significant feature sets for effective thermal comfort transfer learning. We find the combination of age, gender, outdoor environmental features and six factors from the PMV model can lead to the best prediction performance for transfer learning based thermal comfort modelling. 
    \end{itemize}

The remainder of the paper is organized as follows. Section~\ref{sec: relatedwork} presents the related work on transfer learning and thermal comfort modelling. Section~\ref{sec: dataset introduction} introduces the datasets and shows preliminary analysis. Section~\ref{sec:methodogy} demonstrates the proposed thermal comfort modelling framework. Section~\ref{sec:experiment} contains the result of experiments with the detailed experiment settings, comparison with the state-of-the-art algorithms, and improvement analysis with different configurations (e.g., feature combinations, number of hidden layers). Section~\ref{sec:conclusion} concludes the paper and shows the direction in future work.

\section{Related Work}
This section discusses the background on thermal comfort modelling techniques and transfer learning applications especially on sensing data. Then we summarize the data-driven thermal comfort studies in Table \ref{tab:relatedwork} and address the advantages of this research.
 

\label{sec: relatedwork}
\subsection{Thermal Comfort Modelling} 
Fanger's PMV model and de Richard's adaptive model are the most famous knowledge-driven thermal comfort models. The adaptive model \cite{de1998developing} is based on the idea that occupant can adapt to different temperatures in different times and outdoor weather affects indoor comfort. Occupants can achieve their own comfort through personal adjustments such as clothing changes or window adjustments. Clear et al. \cite{clear2013understanding} explored how adaptive thermal comfort could be supported by new ubiquitous computing technologies. They addressed that IoT sensing technologies can help build a more sustainable environment where people are more active in maintaining and pursuing their thermal comfort, which is less energy-intensive and less tightly controlled.

Recently, data-driven thermal comfort modelling becomes more and more popular and a lot of efforts have been made for applying machine learning and deep learning techniques to human thermal comfort modelling. Ran et al. \cite{ranjan2016thermalsense} used \textit{Rotation Forests} to predict thermal comfort with thermographic imaging involving thirty individuals in a UK office building. Similarly, Ghahramani et al. \cite{ghahramani2018towards} used a hidden Markov model (HMM) based learning method to predict personal thermal comfort with infrared thermography of the human face from 10 subjects. Chaudhuri et al. \cite{chaudhuri2018random} established a random forest-based thermal comfort model for different gender groups (6 females and 8 males) using physiological signals (e.g., skin conductance, blood pressure, skin temperature). However, all the thermal comfort prediction models mentioned above need to install additional devices (individual thermal cameras, smart eyeglass, physiological sensors) and may lead to the privacy concern. 

Luo et al. \cite{luo2020comparing} compared to nine widely used machine learning algorithms for thermal sensation prediction using the ASHRAE Comfort Database II. They found that ML-based thermal sensation prediction models generally have higher accuracy than traditional PMV models and Random Forest has the best performance compared with other ML algorithms. They also addressed the importance of tuning parameters and selecting input features for machine learning models. 

\begin{table*}
 \centering
   \footnotesize
    \setlength\tabcolsep{3pt}
\caption{Related works for the data-driven thermal comfort modelling}
\label{tab:relatedwork}
  \centering
\begin{tabular}{@{}llllllll@{}}
\toprule
\textbf{Reference}         & \textbf{Date} & \textbf{Dataset}                                                                                 & \textbf{Data source} & \textbf{Location}  & \textbf{Time} & \textbf{Approach}                                                          & \textbf{Key findings}                                                                                                                                                                                                                                                                    \\ \midrule
Hu et al. \cite{hu2018itcm}         & 2018          & \begin{tabular}[c]{@{}l@{}}Self-collected sensing \\ and wearable data\end{tabular}              & Field study          & Singapore          & 3 weeks       & \begin{tabular}[c]{@{}l@{}}MLP neural \\ network\end{tabular}              & \begin{tabular}[c]{@{}l@{}}They built an intelligent thermal comfort management\\ system and implemented a MLP network using \\ sensing and wearable data.\end{tabular}                                                                                                                  \\ \hline
Zhang et al. \cite{zhang2018thermal}      & 2018          & \begin{tabular}[c]{@{}l@{}}Medium US Office \\ dataset\end{tabular}                              & Field study          & Philadelphia, USA  & 1 year        & \begin{tabular}[c]{@{}l@{}}Fine-grained\\ MLP network\end{tabular}         & \begin{tabular}[c]{@{}l@{}}They proposed a fine-grained MLP network to model\\ the relationship between controlling building \\ operations and thermal comfort factors.\end{tabular}                                                                                                     \\\hline
Ran et al. \cite{ranjan2016thermalsense}        & 2016          & \begin{tabular}[c]{@{}l@{}}Self-collected thermal\\ images\end{tabular}                          & Lab study            & UK                 & 5 weeks       & Rotation forest                                                            & \begin{tabular}[c]{@{}l@{}}They dynamically predicted thermal comfort by using\\ thermographic imaging information.\end{tabular}                                                                                                                                                         \\\hline
Ghahramani et al. \cite{ghahramani2018towards} & 2018          & \begin{tabular}[c]{@{}l@{}}Self-collected sensing\\ data from eyeglass\end{tabular}              & Lab study            & Los Angeles, USA   & 4 days        & \begin{tabular}[c]{@{}l@{}}Hidden markov \\ model (HMM)\end{tabular}       & \begin{tabular}[c]{@{}l@{}}They presented a HMM based learning method to \\ predict thermal comfort with infrared thermography\\ of the human face.\end{tabular}                                                                                                                         \\\hline
Chaudhuri et al. \cite{chaudhuri2018random}  & 2018          & \begin{tabular}[c]{@{}l@{}}Self-collected \\ physiological sensing\\ data\end{tabular}           & Lab study            & Singapore          & 2 months      & Random forest                                                              & \begin{tabular}[c]{@{}l@{}}They established a random forest-based thermal \\ comfort model for different gender groups using \\ physiological signals.\end{tabular}                                                                                                                      \\\hline
Luo et al. \cite{luo2020comparing}        & 2020          & \begin{tabular}[c]{@{}l@{}}ASHRAE comfort \\ database\end{tabular}                               & Field study          & Multiple locations & N/A           & \begin{tabular}[c]{@{}l@{}}Multiple machine \\ learing models\end{tabular} & \begin{tabular}[c]{@{}l@{}}They compared nine widely used machine learning \\ algorithms for thermal sensation prediction and found\\ that ML-based models have higher accuracy than \\ traditional PMV models.\end{tabular}                                                             \\\hline
Ferreira et al. \cite{ferreira2012neural}  & 2012          & \begin{tabular}[c]{@{}l@{}}Self-collected sensing\\ data\end{tabular}                            & Field study          & Portugal           & 8 days        & \begin{tabular}[c]{@{}l@{}}Multiple neural \\ networks\end{tabular}        & \begin{tabular}[c]{@{}l@{}}They controlled an HVAC system to achieve thermal \\ comfort and energy savings, and investigated several\\ neural network models to calculate the PMV index.\end{tabular}                                                                                    \\\hline
Hu et al. \cite{hu2019heterogeneous}         & 2019          & \begin{tabular}[c]{@{}l@{}}RP-884 database, self-\\ collected sensor data\end{tabular}           & Lab study            & Singapore          & 4 months      & \begin{tabular}[c]{@{}l@{}}MLP neural \\ networks\end{tabular}             & \begin{tabular}[c]{@{}l@{}}They adopted transfer learning for thermal comfort \\ modelling on the lab study and transferring data from\\ data all over the world, but did not consider differences\\ in thermal environments in different climate zones.\end{tabular}                             \\\hline  
This work                   & N/A           & \begin{tabular}[c]{@{}l@{}}RP-884, The Scales\\ Project, Medium US\\ Office dataset\end{tabular} & Field study          & Multiple locations & 1 year        & \begin{tabular}[c]{@{}l@{}}MLP neural \\ networks\end{tabular}             & \begin{tabular}[c]{@{}l@{}}We proposed the transfer learning-based thermal \\ comfort model for the target dataset collected in \\ the wild, and found that transfering data from the \\ same climate zone can benefit the thermal comfort \\ modelling in target building.\end{tabular} \\ \bottomrule
\end{tabular}
\end{table*}

In recent years, the use of the artificial neural network in thermal comfort modelling in buildings has been increasing significantly. Ferreira et al. \cite{ferreira2012neural} addressed the problem of controlling an HVAC system with the purpose of achieving a desired thermal comfort level and energy savings. Several neural network models have been investigated to calculate the PMV index for the model based predictive control, which provide good coverage of the thermal sensation scale. Hu et al. \cite{hu2018itcm} built an intelligent thermal comfort management system using sensing data and wearable data. They implemented a black-box MLP neural network for thermal comfort modelling, which displayed better prediction performance than the PMV model, traditional white-box machine learning models (e.g., Naïve Bayes, k-Nearest Neighbor and Decision Tree) and classical black-box machine learning models (e.g., Support Vector Machine, Random Forest).

Compared to most previous research using coarse-grained architecture (link input attributes and thermal comfort score directly), Zhang et al. \cite{zhang2018thermal} used the MLP neural network to model the relationship between controlling building operations and thermal comfort factors, their proposed fine-grained DNN approach for thermal comfort modelling outperforms the coarse-grained modelling and the other popular machine learning algorithms. 

\subsection{Transfer Learning Applications}
Though great contributions have been made for improving the prediction accuracy of thermal comfort through various machine learning techniques, there still exists a main bottleneck for data-driven thermal comfort modelling - the accessibility of sufficient thermal comfort data. Transfer learning allows researchers to learn an accurate model using only a tiny amount of new data and a large amount of data from a previous task \cite{dai2009eigentransfer}. 

The transfer learning technique has been applied to many real-world applications involving image/video classification, natural language processing (NLP), recommendation systems, etc. For instance, transfer learning technique has been used for the children's Automatic Speech Recognition (ASR) task \cite{shivakumar2018transfer}. Researchers learn from adult's models to children's models through a Deep Neural Network (DNN) framework. They investigated the transfer learning techniques between adult and children ASR systems in acoustic variability  (layers near input) and pronunciation variability (layers near output), and updated both the top-most and bottom-most layers and kept the rest of the layers fixed. 

Some existing work has focused on transfer learning on the sensor data. Wang et al. \cite{wang2018stratified} proposed a transfer learning based framework for cross-domain activity recognition. Firstly, they used the majority voting technique to obtain the pseudo-label of the target domain. Intra-class knowledge transfer was interactively performed to convert two domains into the same feature subspace. After that, labels for the target
domain can be ignored by the second annotation. Ye et al. \cite{ye2018slearn} learned human activity labels by leveraging annotations across multiple datasets with the same feature space even though the datasets may have different sensing deployment, sensing technologies and different users.

Recently, a transfer active learning framework was proposed to predict thermal comfort \cite{emil2019transfer}. They considered the thermal comfort prediction as inductive transfer learning where labelled data is available in both source and target domains but users do not have access to all labelled data in target domain. They used parameter transferred from the source domain to target domain. The biggest disadvantage of their method is that they assume the feature spaces in both domains must be same, which is not applicable in the daily life as there may exist unique useful features in the target dataset. 

Similarly, Hu et al. \cite{hu2019heterogeneous} adopted transfer learning for thermal comfort modelling and assumed the feature space of source domain is a subset of that of target domain. They connected the classifiers from source domain and target domain and then built a new classifier to obtain knowledge from source domain, but did not explain why the network structure works well. Besides, they trained the thermal comfort model for the lab study and learn knowledge from data from buildings all over the world in the ASHRAE RP-884dataset, but did not consider the differences of thermal environments in different climate zones.

The data-driven thermal comfort modelling works are summarized in Table ~\ref{tab:relatedwork}. Overall, there are several advantages of our work: (1) we are the first to transfer the knowledge from similar thermal environments (climate zones) to the target building for effective thermal comfort modelling. Most previous research focus on building thermal comfort model for one target building \cite{hu2018itcm,zhang2018thermal,ranjan2016thermalsense,ghahramani2018towards,chaudhuri2018random,luo2020comparing,ferreira2012neural}. Even though a few research \cite{hu2019heterogeneous} started to apply transfer learning for building thermal comfort model, their target dataset is collected from the lab study and does not consider the influences of different climate zones; (2) unlike some research use data collected from the lab study \cite{ranjan2016thermalsense,ghahramani2018towards,chaudhuri2018random,hu2019heterogeneous}, we build the thermal comfort model using data from the field study in both target and source domain, which is much more meaningful in real-world scenarios; (3) compared with some research utilising additional devices (e.g., thermal cameras in \cite{ranjan2016thermalsense}, eyeglass in \cite{ghahramani2018towards}, wristbands in \cite{hu2019heterogeneous}), our research 
is easier and cheaper to be carried out, and better to protect the privacy of occupants.

\section{Data Sets Introduction}
\label{sec: dataset introduction}

In this section, we first introduce the basic information of three popular datasets adopted in this research. Then we conduct the preliminary analysis to understand the variables (environmental and personal factors) and patterns in each dataset.

\subsection{Overview}\label{dataset introduction}
\textit{ASHRAE RP-884 Database} \cite{american2004thermal}. The ASHRAE RP-884 database is one of the most popular public databases for human thermal comfort study, which has been used in numerous previous research \cite{toe2013development,langevin2012relating,lu2019data,von2008correlation}. ASHRAE RP-884 dataset was initially collected to develop De Dear's adaptive model, involving more than 25,000 observations collected from 52 studies and 26 cities over different climate zones all over the world. We adopt this public database as one of the source datasets in our research.

\begin{table*}
\caption{Information for Source Dataset and Target Dataset}
\label{tab:compare}
\begin{tabular}{@{}llll@{}}
\toprule
\textbf{Dataset} & \textbf{ASHRAE RP-884}               & \textbf{The Scales Project} & \textbf{Medium US Office}      \\\midrule
Instances        & 25,623            &  8225& 2,497        \\
Participants   & Unknown  (48\% M, 52\% F)             &   8225  (53\% M, 46\% F)           & 24  (33\% M, 67\% F)                            \\
Indoor AT Range (C)   & 6.2 - 42.7                   &  13.2 - 34.2      & 17.9 - 27.8                    \\
Indoor RH Range (\%)  & 2.0 - 97.8                   &     18.0 - 82.4   & 15.7 - 72.4                    \\
Indoor AV Range (m/s) & 0.01 - 1.71                  &   0.00 - 0.70     & 0.02 - 0.19                    \\
MR Range (Met)  & 0.64 - 6.82                        & NaN  & 1.00 - 6.80                    \\
CL Range (Clo)  & 0.04 - 2.29                        & NaN & 0.21 - 1.73         \\\bottomrule 
\end{tabular}

\end{table*}

\begin{figure}[b]
    \centering
    \includegraphics[width=0.48\textwidth]{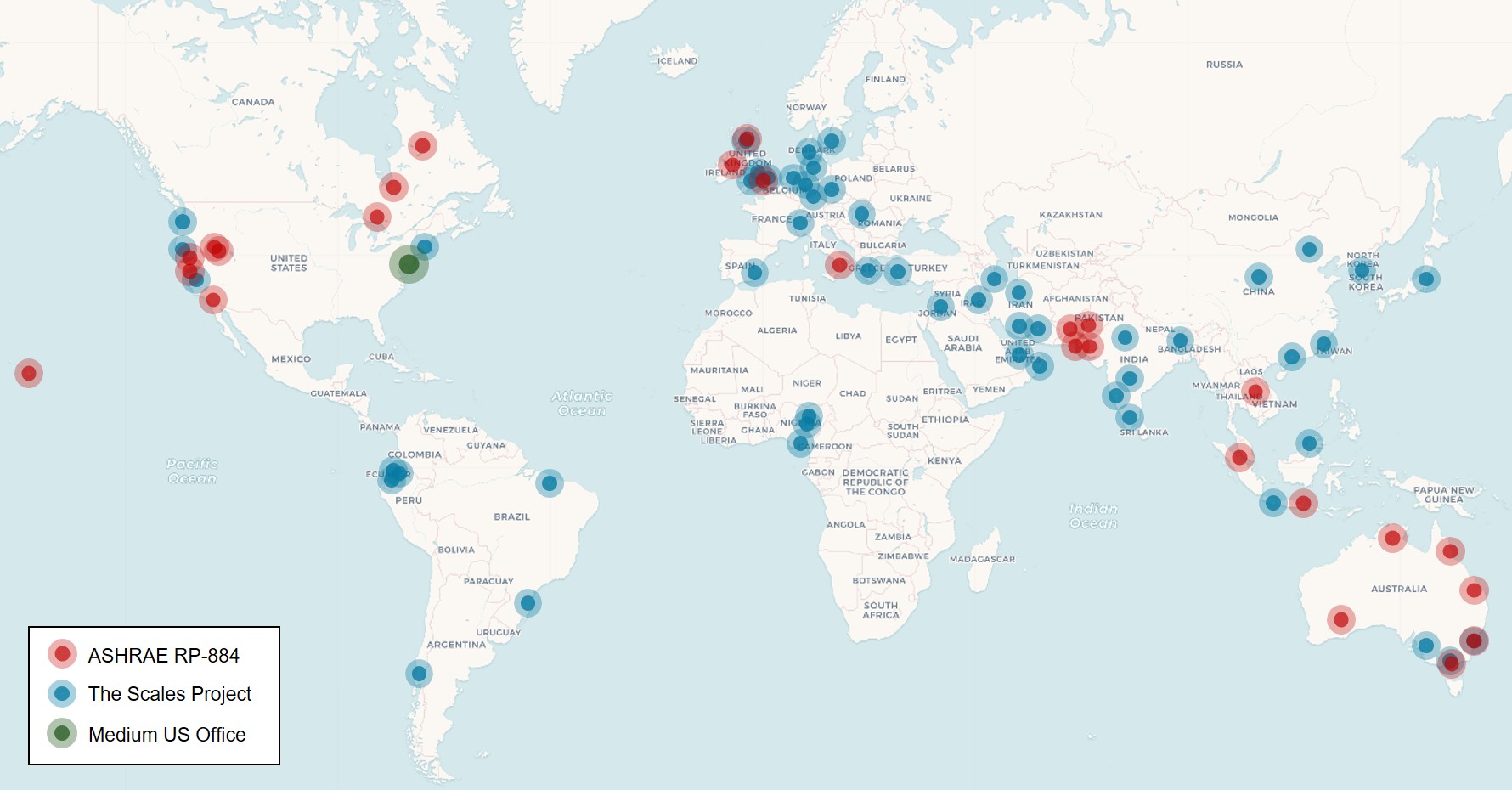}
    \caption{Locations of different studies in ASHRAE RP-884 database, The Scales Project database and Medium US Office dataset}
    \label{fig:citymap}
\end{figure}

\textit{The Scales Project Dataset} \cite{schweiker2019scales}. The Scales Project dataset is published in 2019 which contains thermal comfort responses from 57 cities in 30 countries for 8225 participants. This dataset aims at exploring participants' thermal comfort, thermal sensation, thermal acceptances and to investigate the validity of assumptions regarding the interpretation of responses from the survey. This public dataset is used as one of the source datasets in the research.

%

\textit{Medium US Office Dataset} \cite{langevin2015tracking}. Medium US Office dataset developed by Langevin et al. \cite{langevin2015tracking} is a popular dataset used by recent thermal comfort studies \cite{schweiker2016effect,zhang2018thermal}. It collected data from 24 participants (16 females and 8 males) in the Friends Center Office building in Philadelphia city, USA. Longitudinal thermal comfort surveys are distributed online 3 times daily (morning, mid-day and afternoon) for a continuous 2-week period in each of the four project seasons between July 2012 and August 2013. Data types vary from daily surveys to sensor data including but not limited to the indoor air temperature, air velocity, relative humidity, $CO_2$ concentration and illuminance. This public dataset is used as the target dataset in the research.

The locations of all cities conducting the study are displayed in Figure \ref{fig:citymap}. Red points represent 26 cities in the ASHRAE RP-884 database, blue points mean the 57 cities in the Scales Project dataset, and green point indicates the Philadelphia city in the Medium US Office dataset. In this research, we aim to learn the knowledge from data in cities indicated by red points and blue points, to benefit one building in the Philadelphia (green points). 

Table \ref{tab:compare} shows the basic information for the ASHRAE RP884 database, the Scales Project dataset  and Medium US Office dataset. The first two datasets have different building types (HVAC, naturally ventilated and mixed ventilated) while there is only one HVAC building in Medium US Office dataset. Since the ASHRAE and Scales datasets include different climate zones all over the world, they have wider indoor air temperature ranges than Friends Center building in Medium US Office (17.9$^{\circ}C$-27.8$^{\circ}C$). Different to the ASHRAE RP-884 database and the Scales Project dataset, Medium US Office have much smaller participants.  Besides, the range of indoor relative humidity (Indoor RH), indoor air velocity (Indoor AV), metabolic rate (MR), clothing level (Clo) in Medium US Office is smaller than the ASHRAE dataset. In Medium US Office dataset, we found that 67\% thermal survey responses are from female participants and in the ASHRAE/Scales datasets, the responses from male and female are almost the same.

\subsection{Preliminary Analytics}
\label{subsec: preliminary analtyics}
In this section, we conduct some preliminary analytics between the ASHRAE RP-884 dataset, the Scales Project dataset and Medium US Office dataset. We observe that these three datasets have plenty of similarities, and of course, there are some differences. 

\begin{figure}
  \centering
  \subfigure[ASHRAE RP-884]{\includegraphics[width=0.15\textwidth]{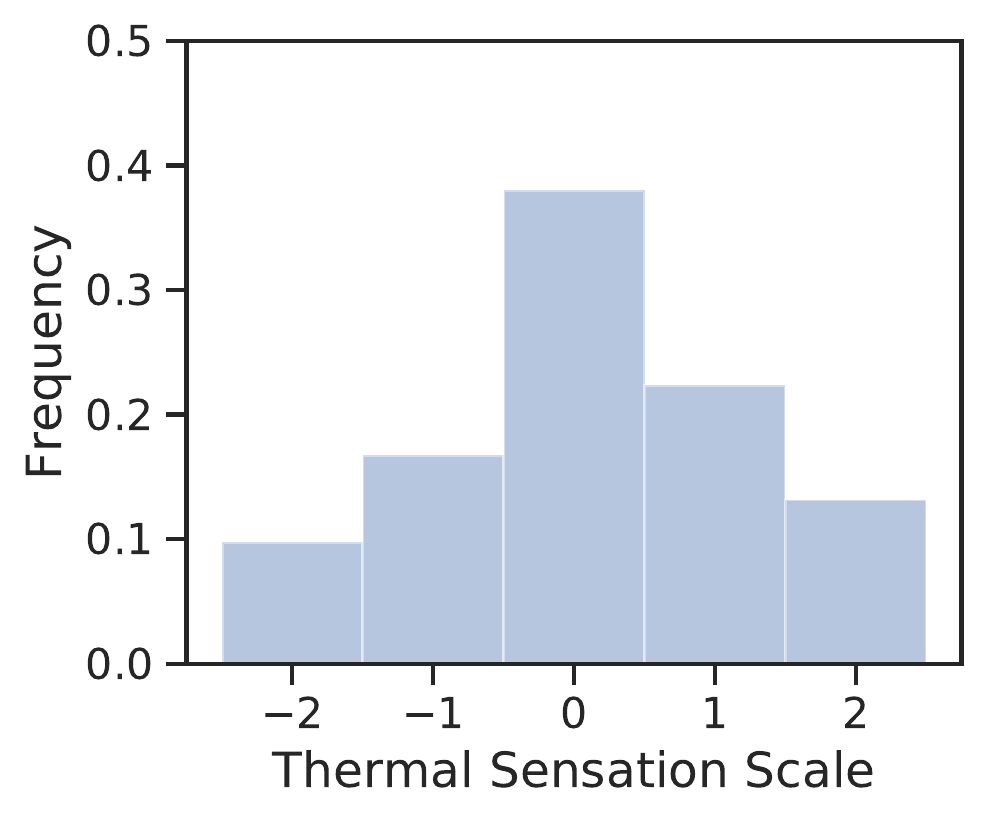}}
  \subfigure[The Scales Project]{ \includegraphics[width=0.15\textwidth]{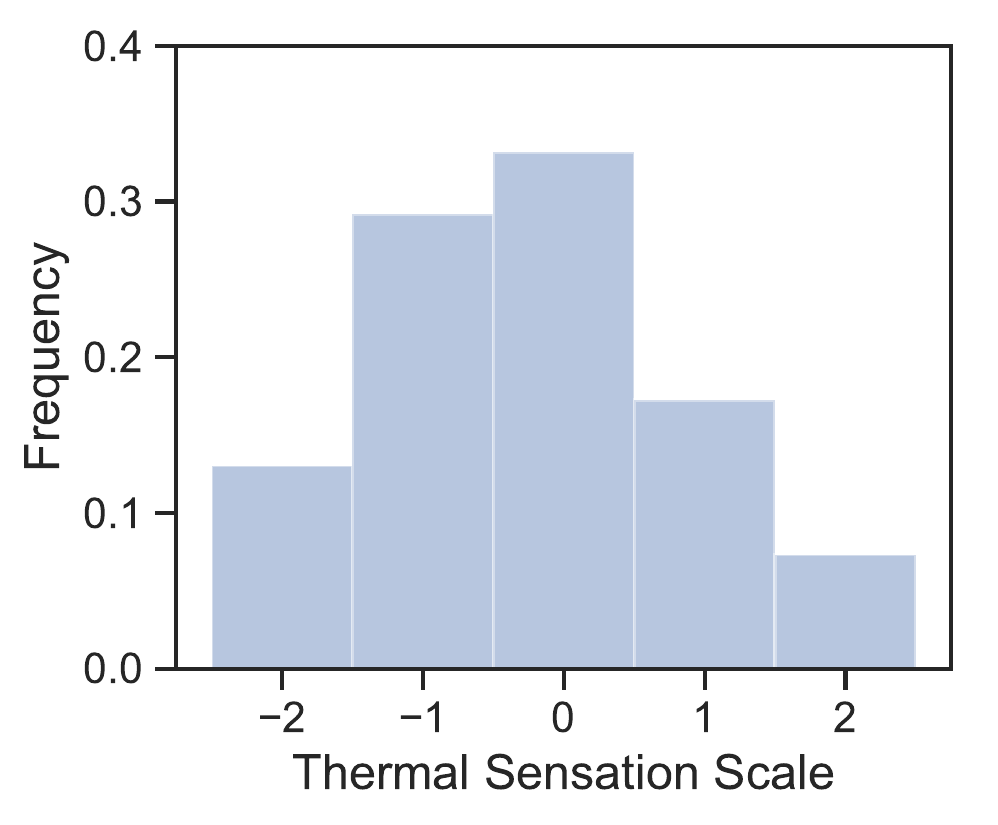}}
  \subfigure[Medium US Office]{ \includegraphics[width=0.15\textwidth]{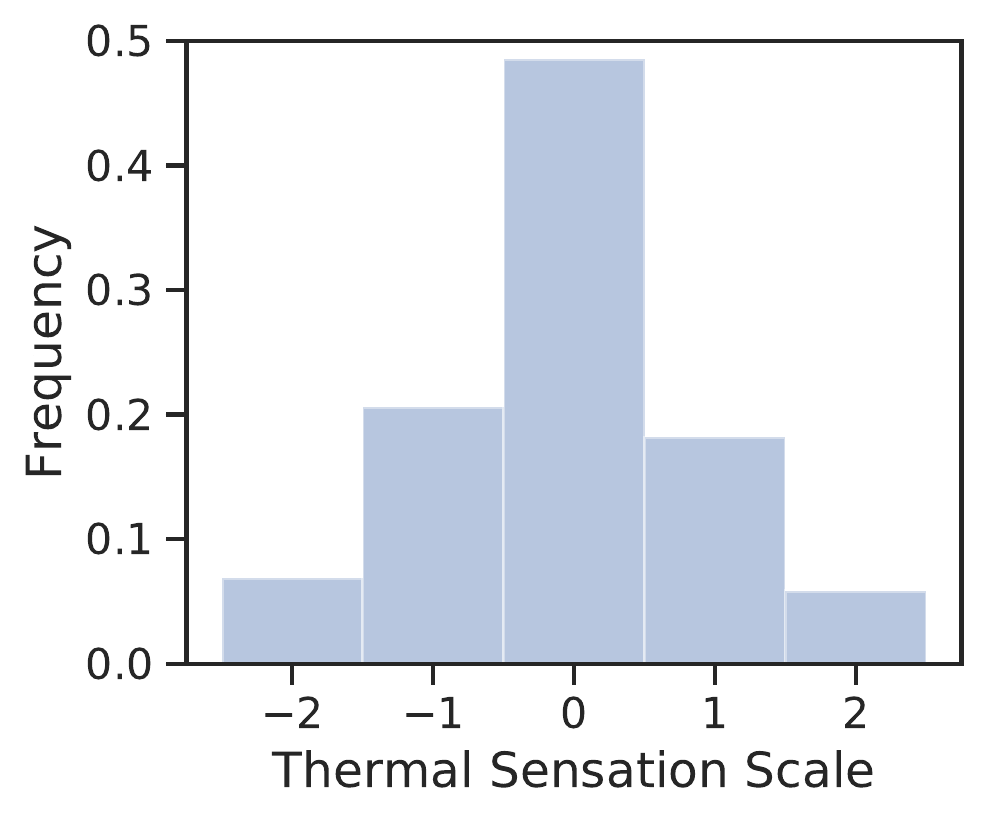}}
  \caption{Distribution of Thermal Sensation over Different Datasets}
\label{fig:dis_sen}
\end{figure}

Figure \ref{fig:dis_sen} shows the distribution of thermal sensation for the ASHRAE RP-884 dataset, the Scales Project and Medium US Office dataset. Since the instances of sensation scale for +3 (Hot) and -3 (cold) is far less than the other instances in both data sets, we merged +3 (hot) and +2 (warm) into one class, and -3 (cold) and -2 (cool) into one class. In the office environment, indoor environmental factors such as temperature are generally maintained at a relatively comfortable level (17.9$^{\circ}C$-27.8$^{\circ}C$ in Medium US dataset), people can also choose to adjust their clothing level and behaviour (e.g., open the heater, have hot drinks) in case of too cold or too hot. 

\begin{figure}[b]
    \centering
    \includegraphics[width=0.3\textwidth]{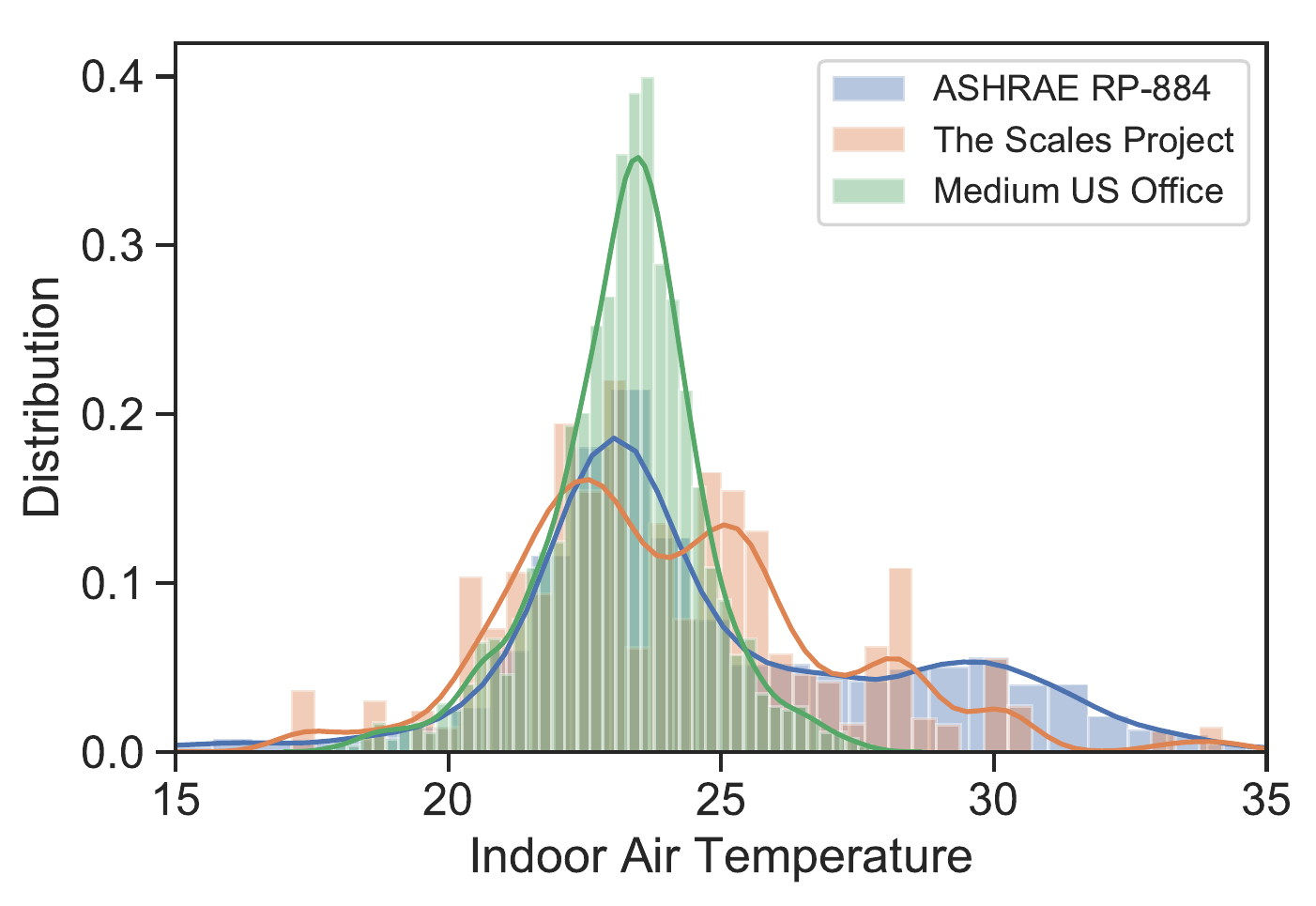}
    \caption{Distribution of indoor air temperature over different domains}
    \label{fig:dis_temp}
\end{figure}

Although the regression model is effective in many times-series problems \cite{shao2018parking, 
nan2019Predicting}, classification method still dominate the thermal comfort area. Therefore, in this paper, we choose the classifiers rather than regressors for effective thermal comfort prediction. Besides, based on the previous discussion, thermal sensation scales are classified into 5 categories (i.e., cold or cool, slightly cool, neutral, slightly warm, hot or warm). 

For the above three datasets, they have similar thermal sensation distributions and occupants feel neutral towards the thermal environment in most time. We can observe that there are more responses for feeling slightly warm or cool than feeling warm/cool or hot/cold, which accords with our thermal comfort feelings in daily life. Meanwhile, the thermal sensation distributions in the ASHRAE dataset and the Scales Project dataset are more uniform than the distribution of the Medium US Office dataset. This is because the ASHRAE dataset and the Scales Project dataset consist of a variety of data from different climate zones all over the world while Medium US Office dataset only includes data from one building.

\begin{figure}
  \centering
  \subfigure[ASHRAE Dataset\label{fig:box1}]{\includegraphics[width=0.15\textwidth]{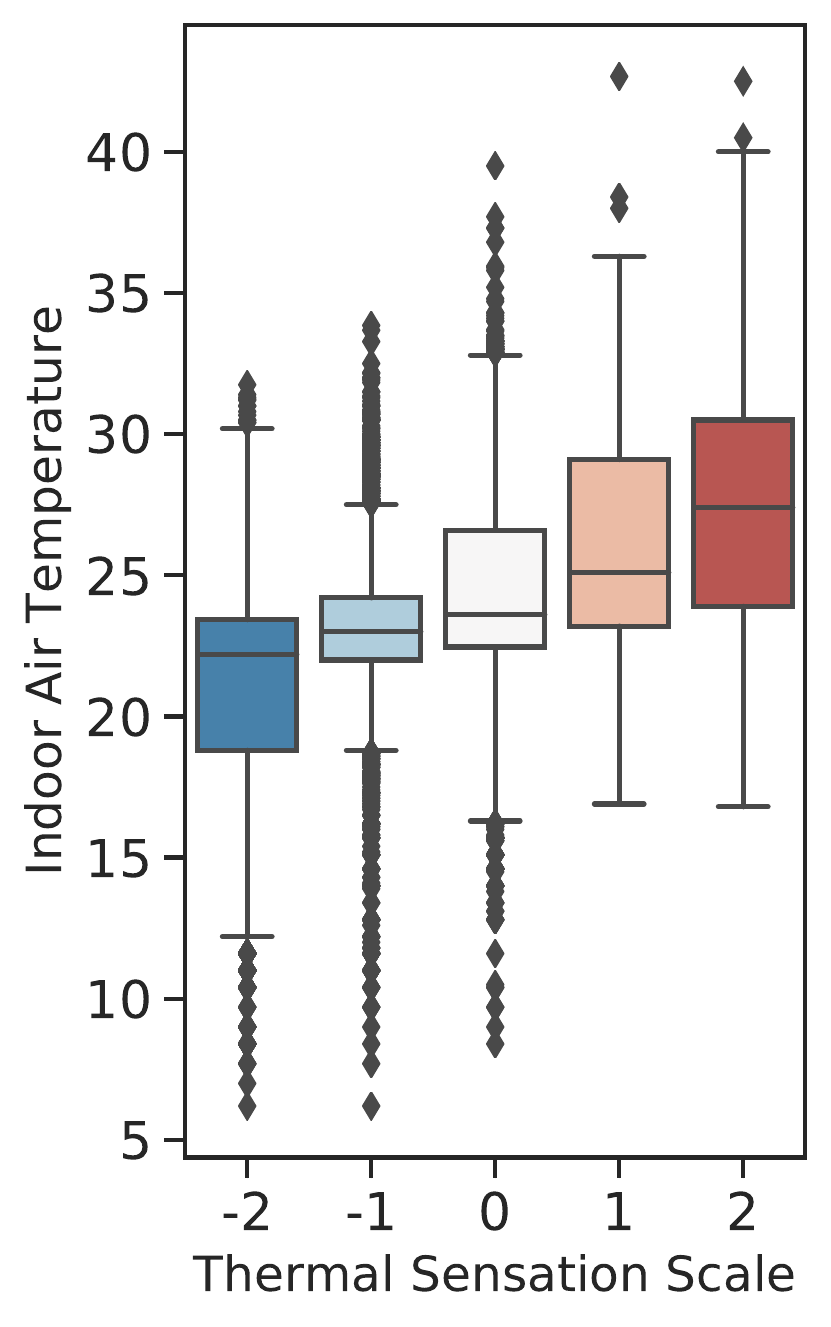}}
  \subfigure[The Scales Project Dataset\label{fig:box_scales}]{\includegraphics[width=0.15\textwidth]{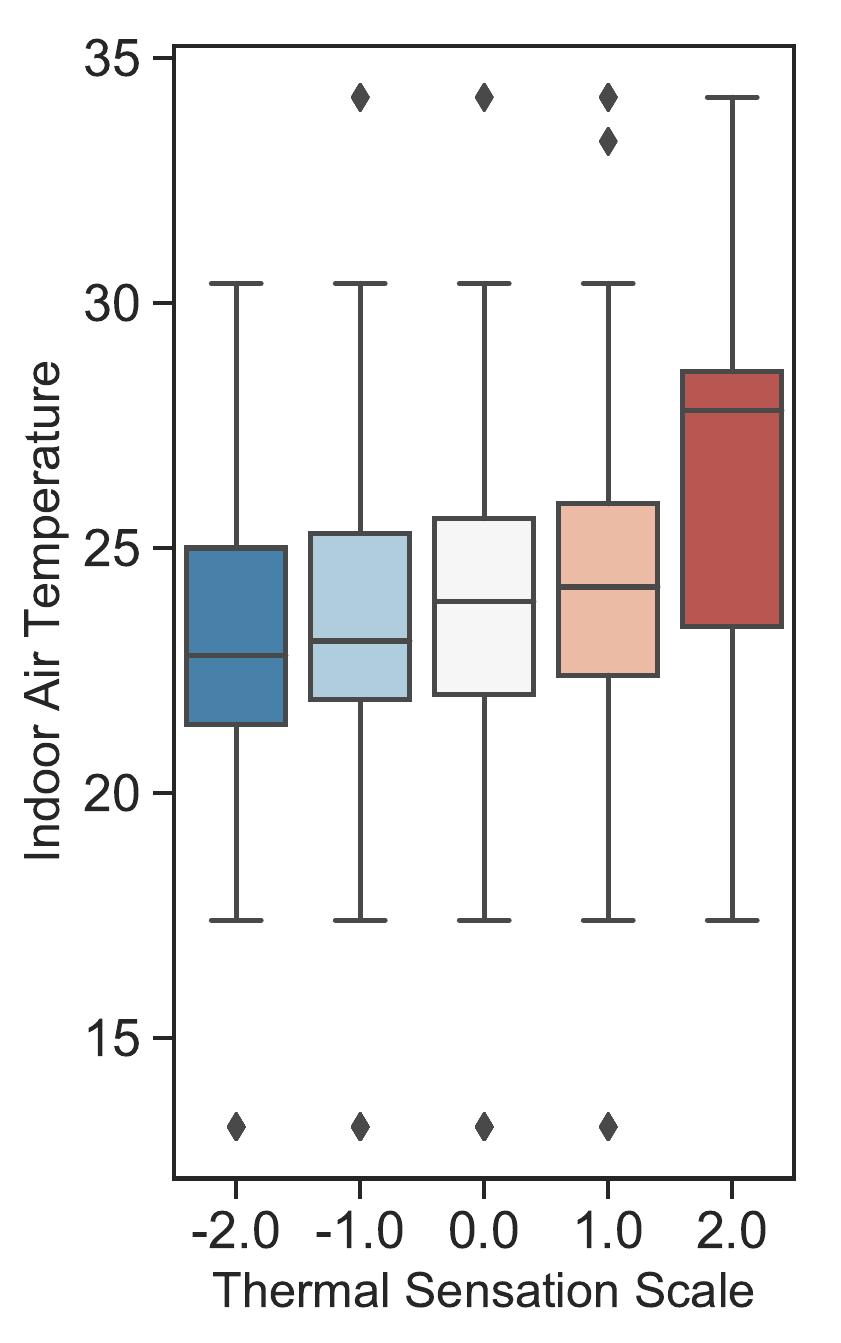}}
  \subfigure[Medium US Office dataset\label{fig:box2}]{\includegraphics[width=0.15\textwidth]{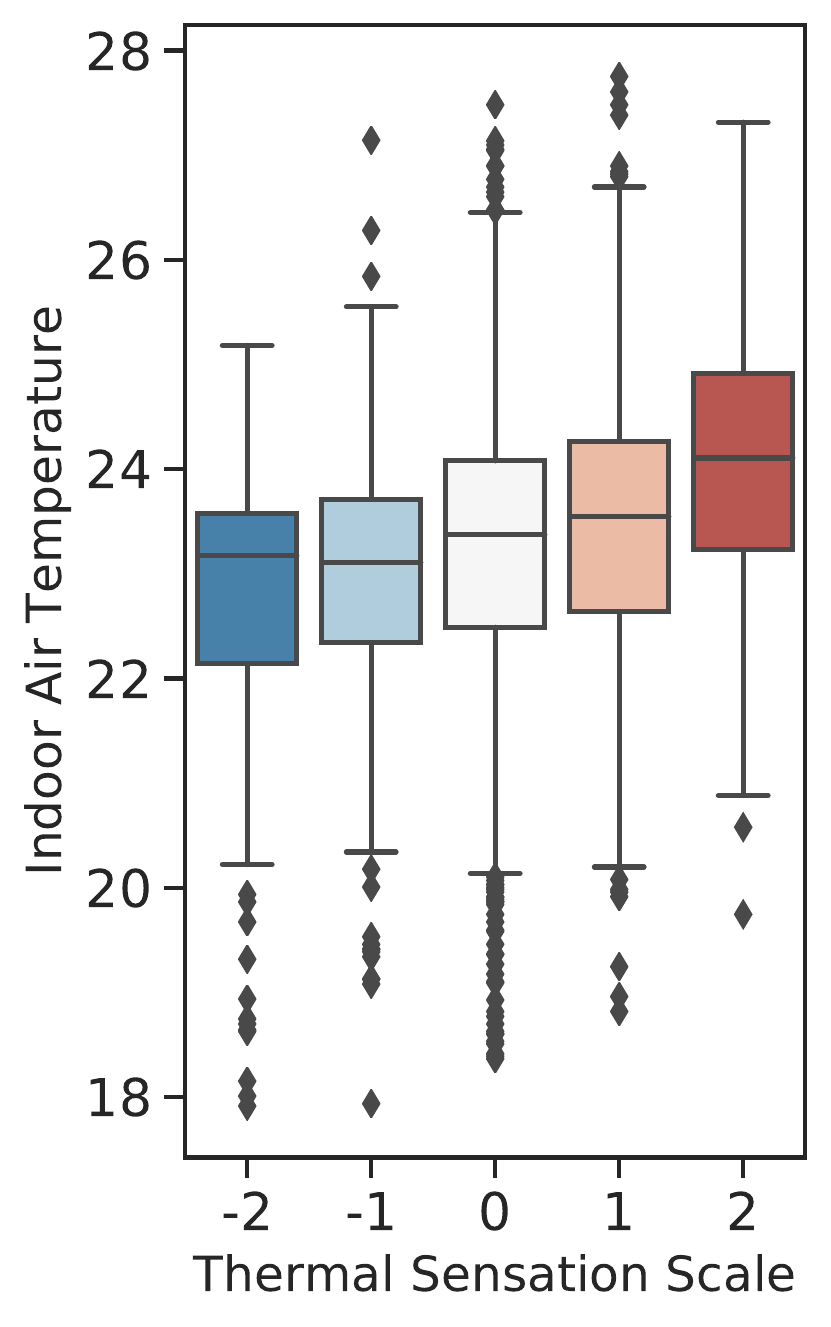}}
  \caption{The boxplots of thermal sensation and indoor temperature}
  \label{fig:box}
\end{figure}
Indoor air temperature is one of the most significant factors affecting occupant's thermal feelings. Figure \ref{fig:dis_temp} shows the distribution of indoor air temperature for the three datasets, most temperature values are between 22$^{\circ}C$-24$^{\circ}C$. However, there are also some differences between these three distributions. The ASHRAE and the Scales Project dataset has higher indoor air temperature due to the fact that some thermal sensations responses are from the hot climate area. On the contrary, in Medium US Office dataset, the indoor temperature distribution seems to be centered around 20$^{\circ}C$ to 27$^{\circ}C$.

From Figure \ref{fig:box}, we can see the relationship between the indoor air temperature and thermal sensation scale. Usually, a higher indoor air temperature indicates a higher thermal sensation scale for all three datasets. Interestingly, in Medium US Office dataset, the average indoor air temperature for feeling cold or cool is a bit higher than that for feeling slightly cool. This phenomenon may be due to too few subjects (24 participants in total) in the Medium US Office dataset. Also, the other factors such as relative humidity, age, gender, outdoor weather will affect the thermal sensation. That's the reason why we try to use as many features to build a more accurate and robust thermal comfort prediction model.

\begin{figure}
  \centering
  \subfigure[ASHRAE Dataset\label{fig:met_1}]{\includegraphics[width=0.22\textwidth]{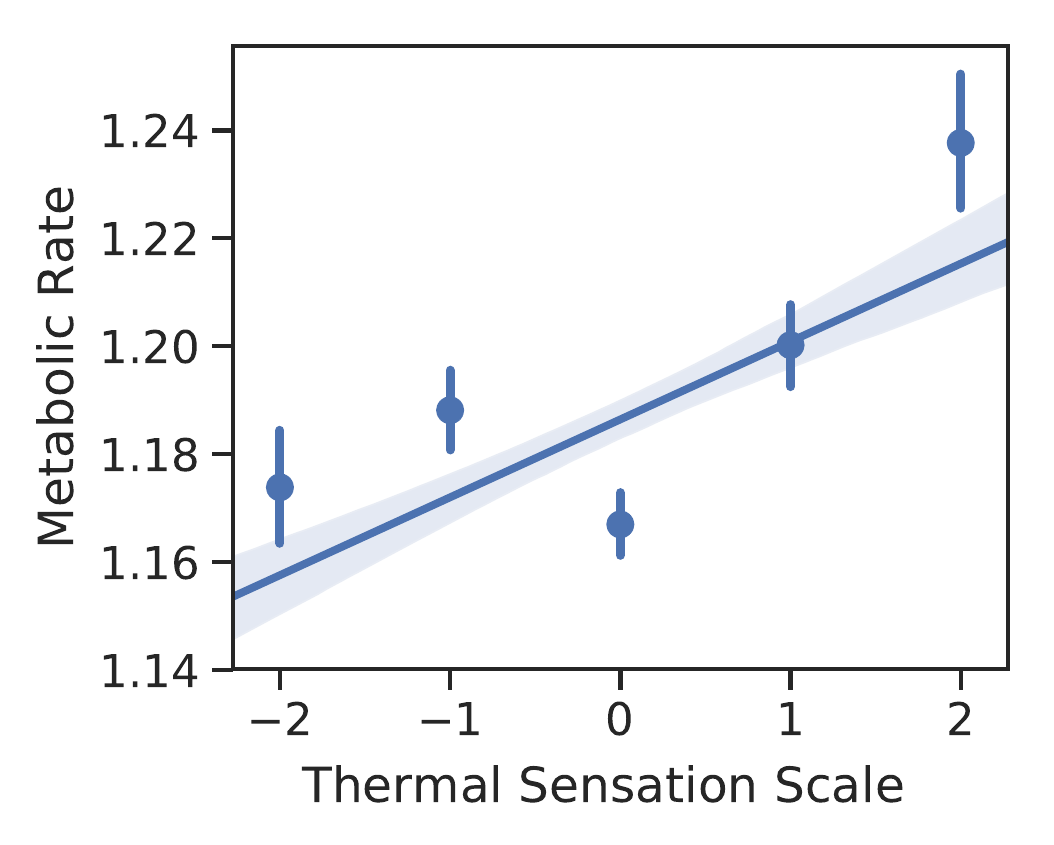}}
  \subfigure[Medium US Office dataset\label{fig:met_2}]{\includegraphics[width=0.21\textwidth]{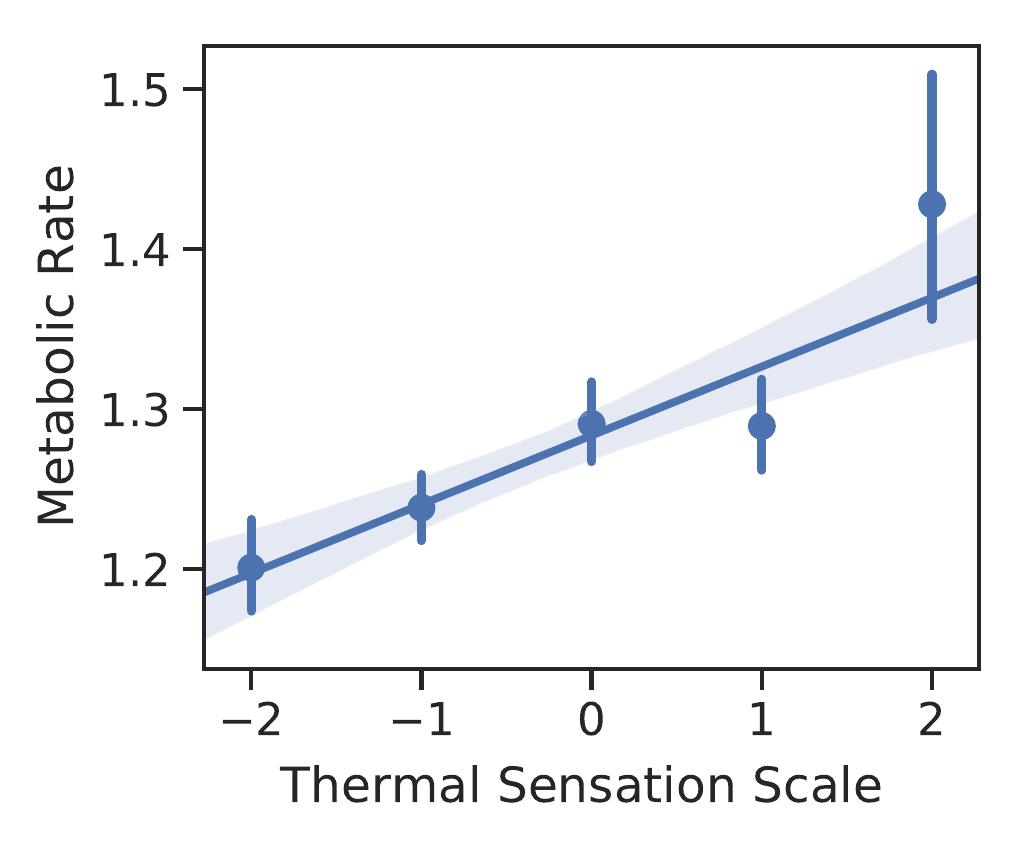}}
  \caption{The boxplots of thermal sensation and metabolic rate}
  \label{fig:met}
\end{figure}

Figure \ref{fig:met} indicates that, in ASHRAE and Medium US Office datasets, the metabolic rate has a positive relationship with the thermal sensation scale. Although they have different average values and confidence intervals for metabolic rate, the trend of how thermal sensation scale changing over metabolic rate seems the same.

From the above analysis, there are observable differences between the ASHRAE, the Scales Project and Medium US Office datasets. One of the reason is that buildings in these three datasets are located in various climate zones, where the climate variability can lead to a different working environment, occupant cognition and behaviour, therefore affecting occupants' thermal sensation in different buildings. Considered the three datasets share lots of similarities in occupant thermal comfort, and the number of instances in the target dataset is very limited, we then explore to infer occupants' thermal comfort by learning from multiple buildings in the same climate zone with similar climate conditions. We will then introduce the proposed thermal comfort modelling in Section~\ref{sec:framework}.



\section{Methodology}
\label{sec:methodogy}
In this section, we introduce the proposed thermal comfort transfer learning framework. Firstly, we introduce the problem definition and then discuss the selected features in the source datasets and target dataset. After that, we demonstrate the methods of dealing with imbalanced thermal comfort dataset. Lastly, we explain the proposed thermal comfort modelling framework.

\subsection{Problem Definition}
To learn sensor data from multiple datasets for thermal comfort modelling, some notations need to be defined in this paper. Firstly, we give the definition of a 'task' and a 'domain'. A domain $\mathcal{D}$ can be represented as $\mathcal{D} = \left\{{\mathcal{X},P(X)}\right\}$, which contains two parts: the feature space $\mathcal{X}$ and the marginal probability distribution $P(X)$, where X = $ \left\{x_1,x_2,...,x_n\right\}\in\mathcal{X} $. The task $\mathcal{T}$ can be represented as $\mathcal{T} = \left \{ y, f(\cdot ) \right \}$, which contains two components: the label space $y$ and a target prediction function $f(\cdot)$. $f(\cdot)$ can not be observed but can be learnt from the training data, which could also be considered as a conditional function $P(y|x)$.

In our research, we aim to transfer the knowledge from the source domain (RP-884 and the Scales Project datasets) to benefit the thermal comfort prediction in the target domain (Medium US Office dataset). Although both domains have different features, they share several common features such as indoor air temperature, indoor relative humidity, indoor air velocity, indoor mean radiant temperature, clothing level, metabolic rate, occupants' age and gender. Therefore, predicting thermal comfort falls under \textit{Transductive transfer learning} \cite{arnold2007comparative}, which can be formally defined as: Given a source domain $\mathcal{D}_s$ and the corresponding learning task ${\mathcal{T}}_s$, a target domain $\mathcal{D}_t$ and the corresponding learning task ${\mathcal{T}}_t$, we aims to improve the performance of the prediction function $f(\cdot)_t$ in $\mathcal{T}_t$, by discover the knowledge from $\mathcal{D}_s$ and ${\mathcal{T}}_s$, where $\mathcal{D}_s \neq \mathcal{D}_t$ and $\mathcal{T}_s = \mathcal{T}_t$.

 \begin{table*}
\caption{Selected Features in Medium US Office Dataset}
\label{tab:features_us}
\centering

\begin{tabular}{@{}cllll@{}}
\toprule
\textbf{Category}                                                & \textbf{Data Source}                & \textbf{Feature Name}    & \textbf{Description}               & \textbf{Units} \\ \midrule
\multirow{4}{*}{\textit{Indoor}} & \multirow{4}{*}{HOBO Datalogger (15 mins)}    & Indoor\_AT      & Indoor temperature         & $ ^{\circ}C $       \\
                                             &                                     & Indoor\_RH       & Indoor relative humidity   & $\%$             \\
                                           &                                     & Indoor\_AV             & Indoor air velocity        & $m/s$            \\
                                           &                                     & Indoor\_AMRT & Indoor radiant temperature    & $ ^{\circ}C $       \\
                                  [0.1em] \hline \\[-1em]
\multirow{2}{*}{\textit{Outdoor}}                    & \multirow{2}{*}{Weather Analytics (15 mins)}  & Outdoor\_AT     & Outdoor temperature        & $^{\circ}C $        \\     &                                     & Outdoor\_RH        & Outdoor  humidity           & $\%$             \\                               [0.1em]\hline \\[-1em]
\multirow{4}{*}{\textit{Personal}}                 & \multirow{2}{*}{Daily Survey (3 times/day)}      & CL      & Clothing insulation  & \textit{clo}            \\
                                                                 &                                     & MR         & Metabolic rate       & \textit{Met}            \\ [0.1em]\cline{2-5} &  
                                                               \multirow{2}{*}{Background Survey (once)}                                                          & Age           & Participant's age      & \textit{Years}            \\    
                                                              && Gender           & Participant's gender    & \textit{}            \\    
\cmidrule(l){1-5} 
\end{tabular}

\end{table*}

\begin{figure}[b]
    \centering
    \includegraphics[width=0.40\textwidth]{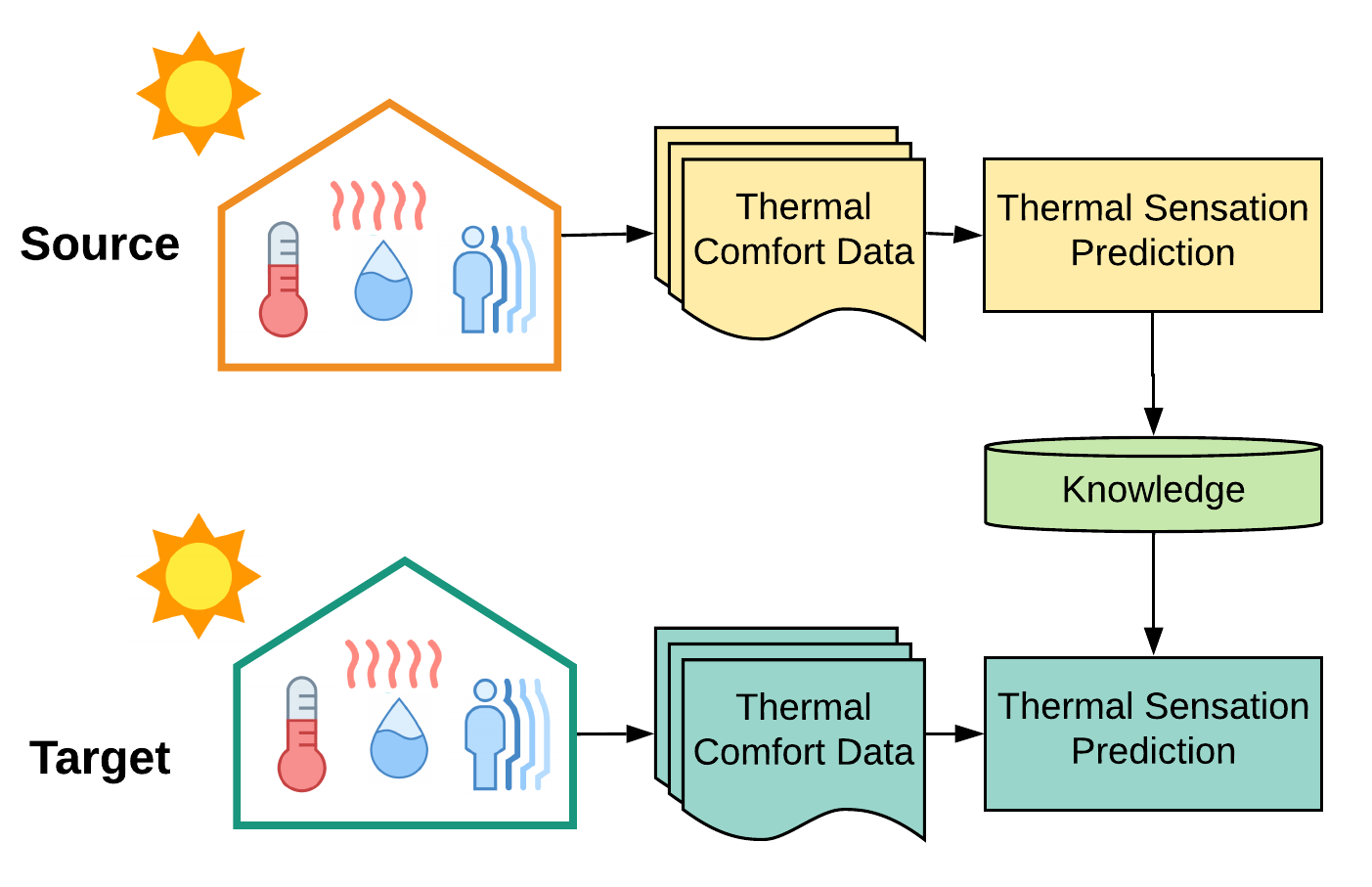}
    \caption{Thermal comfort transfer learning system}
    \label{fig:illustration}
\end{figure}

Figure \ref{fig:illustration} shows the thermal comfort transfer learning system, in which we could use the transfer learning method to learn knowledge from the source datasets and benefit the target dataset in a specified city.
\subsection{Feature Selection}
\label{subsec:featureselection}
Human thermal sensation is influenced by a variety of factors such as time factors \cite{auffenberg2015personalised}, personal information \cite{chaudhuri2018random}, environmental changes  \cite{seppanen1999association}, and culture \cite{indraganti2010effect}. In this research, several  features are chosen for thermal comfort transfer learning based on the following criteria: (1) the features are commonly studied in previous thermal comfort research; (2) the features are
easy to be calculated or collected by passive sensing or self-report responses. In sum, we divide the features into three broad categories: indoor environmental features, outdoor environmental features and personal features. Table \ref{tab:features_us} displays the selected features in Medium US Office dataset.

\textbf{\textit{Indoor Environmental Features}}. Indoor environment affects the occupants' thermal comfort directly and we adopt the following basic indoor environmental features derived from Fanger's PMV model \cite{fanger1970thermal}: air temperature, mean radiant temperature, air velocity and relative humidity for thermal comfort prediction. Air temperature is the average temperature of the air surrounding the occupant around the location and time. Radiant temperature indicates the radiant heat transferred from a surface, and the mean radiant  temperature is affected by the emissivity and temperature of surrounding surfaces, viewing angles, etc. Air velocity means the average speed of air with respect to the direction and time. The relative humidity is the ratio of amount of water vapour in the air to the water vapour that the air can hold at specified pressure the temperature.

\textbf{\textit{Outdoor Environmental Features}}. Outdoor weather conditions can have physiological effects on individuals thermal perception and clothing preference in different seasons \cite{becker2009thermal,zhang2020outdoor}. For instance, in summer people tend to choose lightweight clothing, which will have an influence on their indoor thermal comfort. The most popular measurements of outdoor environment include outdoor air temperature, outdoor humidity, which will also be adopted in this research.


\textbf{\textit{Personal Features}}. Studying personal features is crucial for effective thermal comfort modelling because thermal sensation is a subjective measurement and different individuals perceive the same environment differently. In this research, we select the following personal features: clothing insulation, metabolic rate, age and gender. Clothing insulation has a major impact on thermal comfort level because it affects heat loss and thus the heat balance. Previous research shows the relationship between age and thermal sensation \cite{indraganti2015thermal,indraganti2010effect}. Besides, Sami et al. \cite{karjalainen2007gender} found a significant gender difference in thermal comfort: females tend to prefer a higher room temperature than males and feel both uncomfortably hot and uncomfortably cold more often than males. Hence, gender and age are considered as the features for thermal comfort modelling. 

 
The features in a source domain can be considered as a subset in target domain. The ASHRAE dataset shares eight features with the Medium US Office dataset while the Scales Project dataset only shares six features with target dataset. Although there are various other features in the three datasets above such as occupant behaviour data (e.g., adjust heaters/ curtains/ thermostats) and background survey (e.g., acceptable temperature), we just simplify the thermal comfort prediction and therefore do not show the other features. 


\subsection{Imbalance Class Distribution}
As thermal sensation scale are 5-point values, we regard thermal comfort prediction as a classification task. Fig.~\ref{fig:dis_sen} shows the distributions of the ASHRAE RP-884, the Scales Project and Medium US Office datasets. It is clear that the three distributions are imbalanced and the number of thermal sensation instance for -1 (cool) to 1 (warm) far exceeds the other instances. To train a fair classifier, we must deal with this class imbalance issue in thermal comfort data. Take the binary classification as an example. If class $M$ is 95\% and class $N$ is 5\% in the dataset, we can simply reach an accuracy of 95\% by predicting class $M$ each time, which contributes a useless classifier for our purpose. In this research, we assume the survey responses are 'correct'. Although there may exist some biases (e.g., rating bias, anchoring bias, social-desirability bias) in self-report data, we will not discuss them in this paper. 

To deal with the imbalanced dataset, oversampling and undersampling are efficient techniques to adjust the class distribution of the data set. Under-sampling (e.g., \textit{Clustering}, \textit{Edited Nearest Neighbours} \cite{wilson1972asymptotic}, \textit{Tomek Links} \cite{tomek1976two}) can balance the dataset by reducing the size of the majority class. However, under-sampling methods are usually used when we have sufficient data. Oversampling (e.g., \textit{Synthetic Minority Oversampling Technique} \cite{chawla2002smote}, \textit{Adaptive Synthetic Sampling} \cite{he2008adasyn}) aims to balance the dataset by increasing the number of minority classes, which can be applied when the data is insufficient. 

\textit{Generative Adversarial Network} (GAN) has got successful applications in various fields which can learn the probability distribution of a dataset and synthesize samples from the distribution. GAN uses a generator $G$ to capture the underlying data distribution of a dataset and a discriminator $D$ to estimate the probability that a given sample comes from the original dataset rather than being created by $G$. Some techniques such as \textit{TableGAN} \cite{park2018data}, \textit{TabularGAN} \cite{xu2018synthesizing} have been proposed to handle the imbalance of tabular data. Particularly, Quintana et al. \cite{quintana2019towards} have used \textit{TabularGAN} to synthesize a small thermal comfort dataset. They found that when the size of synthesized data is no larger than the size of real samples, the thermal comfort dataset can achieve similar performance to the real samples.

In the thermal comfort classification problem, labelled thermal comfort responses are usually small in samples. Therefore, in this research, we explore to synthesize survey responses for handling the imbalance of thermal sensation classes. \textit{TabularGAN} \footnote{Python package for TabularGAN: \url{https://pypi.org/project/tgan/}} is used in this research to generate tabular data based on the generative adversarial network.  TabularGAN can learn each column’s marginal distribution by minimizing KL divergence, which is suitable for the thermal comfort classification problem compared with other methods such as \textit{TableGAN}, Edited Nearest Neighbors \cite{wilson1972asymptotic}, SMOTE \cite{chawla2002smote}, etc. The reason why we did not adopt \textit{TableGAN} is that it optimizes the prediction accuracy on synthetic data through minimizing cross entropy loss while \textit{TabularGAN} focuses more about marginal distribution. It learns each column's marginal distribution by minimizing KL divergence, which is more suitable for the thermal comfort classification problem.



\begin{figure}
    \centering
    \includegraphics[width=0.49\textwidth]{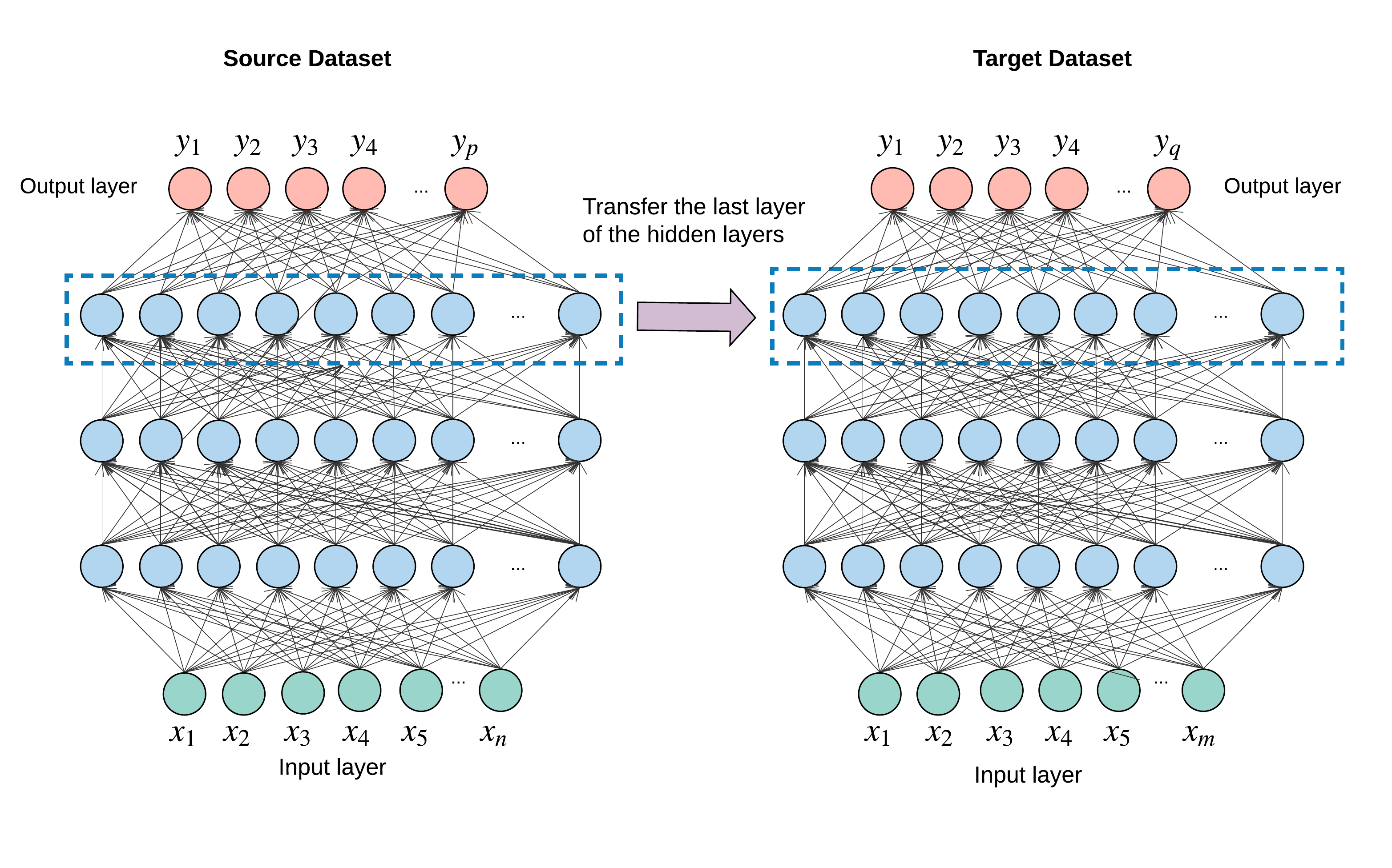}
    \caption{Framework for thermal comfort transfer learning}
    \label{fig:framework}
\end{figure}

\subsection{Thermal Comfort Modelling}\label{sec:framework}

Traditional data-driven thermal comfort modelling is isolated and occurs purely based on specific buildings in the same climate zone. No thermal comfort knowledge is retained which can be transferred from one thermal comfort model to another. Recently, the transfer learning technique has been intensively studied in different applications \cite{emil2019transfer,shivakumar2018transfer}. Transfer learning aims to leverage knowledge from source tasks and then apply them to the target task. There are various transfer learning techniques which can be roughly grouped into three categories: \textit{inductive transfer learning}, \textit{unsupervised transfer learning} and \textit{transductive transfer learning} \cite{pan2009survey}. Inductive transfer learning \cite{daume2006domain} aims to improve performance on the current task after having learned a different but related skill or concept on a previous task. Unsupervised transfer learning \cite{dai2008self} focus on solving unsupervised learning tasks in the target domain such as dimensionality reduction, clustering, and density. Transductive transfer learning aims to utilize the knowledge from the source domain to improve the performance of the prediction task in the target domain.

Transductive transfer learning can exploit the different levels of information captured from different layers in the neural network. Generally, layers close to the input data capture specific characteristics in the dataset while deeper layers could capture information more relevant to the tasks (e.g., object types in image recognition, thermal sensation labels in thermal comfort prediction). The Medium US Office dataset, as described in Section~\ref{dataset introduction}, differs in cities and climate zones from the ASHRAE dataset and the Scales Project dataset. In different climate zones, there are various factors possibly contributing to thermal comfort, e.g., climate characteristics, occupants' recognition and endurance. This motivates us to investigate the transfer learning between the ASHRAE/the Scales Project datasets and Medium US office dataset in climate variability, which is close to the layers near the input. 

We assume the climate variability affects the lower-level neural network only. Therefore, these layers need to be adapted to better represent the Friends Center office building in the target dataset. This can be regarded as retaining the knowledge of higher-level mappings from the source dataset. Hence, we retain the last hidden layer of models in ASHRAE and the Scales Project as shown in Figure \ref{fig:framework}. Then, the thermal comfort neural network will be retrained with Medium US Office dataset until convergence to find optimal parameters for lower hidden layers. 

\section{Experiment}
\label{sec:experiment}

In this section, we conduct experiments on the proposed thermal comfort transfer learning method and compare the performance with the state-of-the-art techniques and different configurations. We aim to address the research questions: \textit{Can we predict occupants’ thermal comfort accurately by learning from multiple
buildings in the same climate zone when we do not have enough data? If so, which features contribute
most for effective thermal comfort transfer learning?}
Specifically, we explore how the numbers of hidden layers and sample size of the training set in the target building affect thermal comfort transfer learning performance.

\subsection{Experiment Setup}

In our research, the source domain (ASHRAE RP-884 dataset and the Scales Project dataset) and the target domain (Medium US Office dataset) share some common features, which includes four indoor environmental variables (air temperature, indoor relative humidity,  mean radiant temperature, indoor air velocity), two environmental variables (air temperature and humidity) and two personal variables (age and gender). At the same time, ASHRAE R[-884 and Medium US Office datasets share another two personal variables (clothing insulation, metabolic rate). The shared features make it possible to transfer knowledge to the target domain from the source domain. 

\textbf{Pre-Processing}. As discussed in Section~\ref{subsec: preliminary analtyics}, we firstly merge the minority classes and reclassify the thermal sensation into five categories. Then, we standardize features by scaling to unity variance for better classification performance. Considering the thermal sensation classes are extremely imbalanced, in order to train a meaningful classifier, the \textit{TabularGAN}  \cite{xu2018synthesizing} technique is applied for synthesizing samples in all the classes except the majority class in the training set. 50\% sample number in each class was synthesized while ensuring that the number of samples per category does not exceed the number of samples in the majority class. 

Take the Medium US Office dataset for example, there are 2497 instances in the original dataset. After removing the null value and categorizing the responses of thermal sensation, there are 1090 responses for 'neutral', 462 responses for 'slightly cool', 408 responses for 'slightly worm', 154 responses for 'cool or cold' and 131 responses for 'warm or hot'. After synthesizing the data using \textit{TabularGAN}, there are 981 responses for 'neutral', 624 responses for 'slightly cool', 551 responses for 'slightly warm', 208 responses for 'cool or cold' and 177 responses for 'warm or hot' in the training set (90\% of the dataset).

\textbf{Architecture}. In this research, we choose the multilayer perception (MLP) neural network as the classifier for the source domain and target domain. Each neural network consists of two hidden layers with 64 neurons in each layer. The Relu function is used as the activation function in hidden layers. Then, the softmax function is applied to the output layer as the activation function. We train the classifier with the categorical cross-entropy loss function and the Adam optimizer with learning rate = 0.001. The batch size is set to 200 and the max epoch have been set to 500. Besides, fixed random seed is chosen for dataset shuffling and training. 

\textbf{Evaluation}. Similar to previous thermal comfort studies \cite{hu2018itcm,ranjan2016thermalsense,chaudhuri2018random,ghahramani2018towards}, accuracy and weighted F1-score are chosen as the performance metrics. Accuracy reflects the overall performance for the thermal comfort model. Since our priority goal is to correctly predict the thermal sensation for as many occupants as possible to achieve the overall thermal comfort/energy saving in the building, accuracy is used as the main evaluation metric in this problem. We also adopt weighted F1-score as the good component to accuracy for capturing performance across imbalanced classes. F1-score takes both false positives and false negatives into account to strike a balance between precision and recall. ‘Weighted-average’ calculate metrics for each class, and find their average weighted by the number of true instances for each class. It takes class imbalance into account compared with the ‘macro-average’ method. Weighted F1-score is helpful for evaluating thermal sensation classifiers as it can take all imbalanced classes into consideration. That is to say, it evaluates the classifiers for different user groups with different thermal sensation levels instead of all occupants globally.


\textbf{Baselines}. For the baseline, three different categories of baselines are selected to compare with our proposed method: Random guess, PMV model and multiple traditional machine learning models. Random guessing generates the sample from the distribution of thermal comfort and regards it as a predicted value. Similar random baselines have been widely used in previous thermal comfort studies such as \cite{abouelenien2017detecting, hu2019heterogeneous}. The PMV model is the most prevalent thermal comfort model all over the world. In the experiment, we will only use the four indoor environmental variables, metabolic rate and clothing insulation to calculate the PMV score ${p}_s$ according to the formula in \cite{da2014spreadsheets} for the target dataset.
Then the thermal sensation class $\mathcal{C}({p}_s)$ is calculated based on the Equation \ref{equ:pmv}.
\begin{equation}
\label{equ:pmv}
\mathcal{C}({p}_s)=\left\{\begin{array}{ll}

-2, & \text { if } {p}_s \leq -1.5 \\ 
-1, & \text { if } -1.5 <  {p}_s \leq -0.5 \\ 
0, & \text { if } -0.5 < {p}_s \leq 0.5 \\ 
1, & \text { if } 0.5 < {p}_s \leq 1.5 \\ 
2, & \text { if } {p}_s \geq 1.5  \\
\end{array}\right.
\end{equation}

For the multiple traditional machine learning models, we choose K-nearest Neighbors \cite{fukunaga1975branch}, Naive Bayes \cite{rish2001empirical}, Support Vector Machine (with Linear, RBF and Polynomial kernel) \cite{suykens1999least}, Decision Tree \cite{safavian1991survey}, Random Forest \cite{liaw2002classification}, AdaBoost \cite{zhao2010research} as baselines. Naive Bayes \cite{rish2001empirical} from Bayes family methods is chosen due to its fast speed and working well with high dimensions. Support Vector Machine \cite{suykens1999least} technique is efficient for handling high dimensional spaces. Different from algorithms like SVM, AdaBoost \cite{zhao2010research} is fast, simple and easy to use with less need for tuning parameters. K-nearest Neighbors \cite{fukunaga1975branch} is a simple method storing all available instances and classifying data instances according to a similarity measure, which has been widely used in the pattern recognition and statistical prediction area. Random Forest \cite{liaw2002classification} is an ensemble learning method for classification operated by building multiple decision trees. It can cope with high-dimensional features and judge the feature importance.

Compared to thes PMV model using six factors for thermal comfort prediction, the multiple machine learning algorithms use ten features as input features (see Table \ref{tab:features_us}). Besides, all the above three baselines build thermal comfort classification model on the Medium US dataset. 



\begin{figure}
    \centering
    \includegraphics[width=0.45\textwidth]{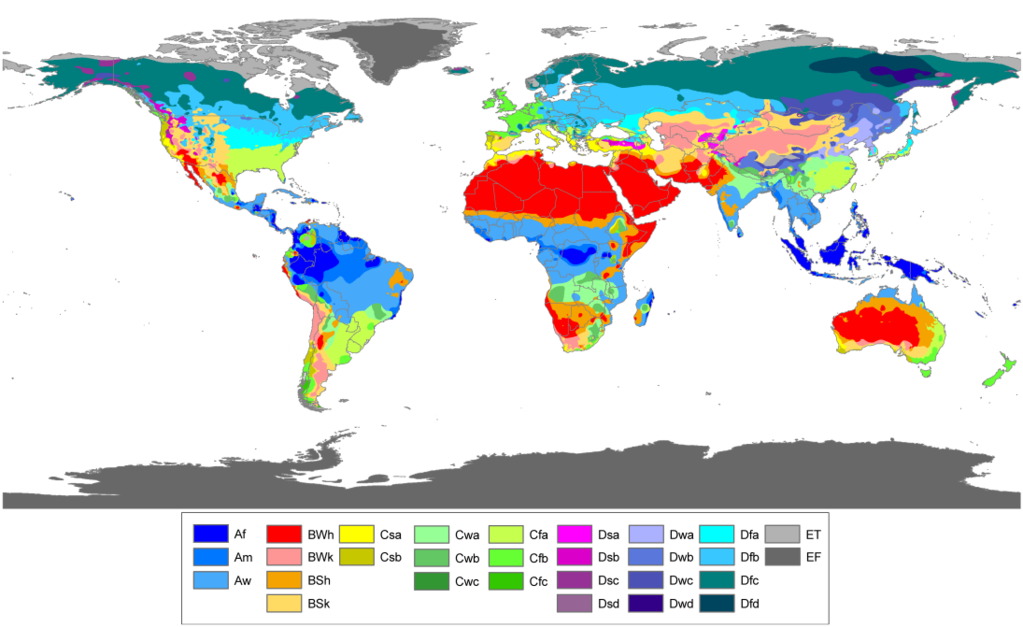}
    \caption{"Köppen World Map High Resolution" by Peel, M. C. et al. \cite{peel2007updated}, licensed under  Creative Commons Attribution-Share Alike 3.0 Unported \cite{creativelisense}, Desaturated from original}
    \label{fig:climate zone}
\end{figure}

\textbf{Cross-validation}. We apply the \textit{$k$-fold cross-validation} \cite{bengio2004no} (k=10) method for effective thermal comfort classification. The advantage of 10-fold cross-validation is to estimate an unbiased generalization performance of the thermal comfort prediction model. In the experiment, the data from the target domain (US Medium Office dataset) is randomly partitioned into 10 folds and each fold serves as the testing data iteratively and the remaining 9 folds are used as training data. The cross-validation process is repeated 10 times and the prediction results (accuracy and weighted F1-score) are averaged to produce a single estimation.  

\begin{table}[b]

\caption{Classification of the ASHRAE RP-884
database for HVAC buildings according to climate}
\label{tab:climaterp}
\begin{tabular}{@{}lll@{}}
\toprule
\textbf{Climate}     & \textbf{Number of cities}                                                                                                                             & \textbf{Instances} \\ \midrule
\textit{Tropical}    & \begin{tabular}[c]{@{}l@{}}5 (Townsville, Jakarta, Darwin,\\  Bankok, Singapore)\end{tabular}                                                                             & 3826                       \\ \hline
\textit{Dry}         & \begin{tabular}[c]{@{}l@{}}6 (Honolulu, Kalgoorlie-Boulder,\\ Karachi,  Quettar, Multan, Peshawar)\end{tabular}                                                           & 3290                       \\ \hline
\textit{Temperate}   & \begin{tabular}[c]{@{}l@{}}12 (Brisbane, Melbourne, Athens, \\ South Wales, Sydney, San Francisco, \\ Merseyside, San Ramon,  Antioch,\\ Auburn, Oxford, Saidu)\end{tabular} & 3512                       \\ \hline
\textit{Continental} & 3 (Ottawa, Montreal, Grand Rapids)                                                                                                                                        & 2808                       \\\hline
\textit{All}         & 26                                                                                                                                                                        & 13436                      \\ \bottomrule
\end{tabular}
\end{table}

\textbf{Climate zone divisions}. We adopt Köppen climate classification updated by Peel et al. \cite{peel2007updated} which is one of the most widely used climate classification systems in the world. As shown in see Figure \ref{fig:climate zone}, the Köppen climate classification divides climates into five main climate zones: A (tropical), B (dry), C (temperate), D (continental), and E (polar). Each large climate zone is then divided into several small subzones based on temperature patterns and seasonal precipitation. All specific climates are assigned a main group of climate zone (the first letter). 

In our study, the target domain (Philadelphia city in the US) belongs to the 'temperate' climate zone. In the source domain, the Scales Project dataset includes 8225 instances from 57 cities in total, and 5411 instances from 32 cities (e.g., Yokohama, Sydney, Cambridge) where located in the 'temperate' climate zone. For the ASHRAE RP-884 database, it consists of 25623 thermal comfort responses from 26 cities in total, where 12 cities (e.g., Berkeley, Athens, Chester) \cite{yong2019exploring} are situated in the same climate zone with Philadelphia city. 

We run the proposed transfer learning based multilayer perceptron (TL-MLP) model and transfer learning based multilayer perceptron from the same climate zones (TL-MLP-C*) model with the ASHRAE database and the Scales Project database as the source domain, the Medium US Office dataset as the target domain. Especially, for both proposed models, we only use the data from HVAC buildings in all datasets. For the TL-MLP-C* model, we use the data from HVAC buildings in the same climate zone as the source domain and Friends Center building as the target domain. 

Besides, we classify the HVAC buildings in ASHRAE RP-884 database to different climates (see Table \ref{tab:climaterp}). It shows that, in ASHRAE RP-884 database, there are 13436 observations from HVAC buildings in total and 3512 observations in the 'temperate' climate zone. Since the Scales Project dataset recorded the Köppen climate and HVAC status information during the data collection, after calculation, there are 4621 observations from HVAC buildings in total and 3245 observations collected from the HVAC buildings located in 'temperate' climate zone.


\subsection{Overall Prediction Result}
\label{sec:compare}

\begin{table}
\centering
\caption{Prediction Performance for Different Algorithms on the Target Dataset}
\label{tab:result}
\begin{tabular}{@{}llll@{}}
\toprule
\textbf{Algorithm} & \textbf{Accuracy(\%)}  & \textbf{F1-score(\%)} \\ \midrule

PMV                & 33.35    (2.40)                            & 32.45          (2.35)      \\ 
Random               & 27.23    (1.30)                           & 29.30         (1.40)      \\ [0.1em]\hline \\[-0.7em]
KNN                & 41.43               (2.95)                & 41.93      (2.85)           \\
SVM (Linear)       & 29.44          (5.19)                    & 30.92           (4.84)      \\
SVM (RBF)       & 37.93             (3.86)             & 40.91               (4.04) \\
SVM (Poly)       & 34.02            (4.59)                 & 37.66                (5.15) \\
Decision Tree      & 43.33      (4.94)                          & 43.34        (4.87)         \\
Random Forest      & \textbf{51.41}    (3.03)                           & 52.93      (3.69)           \\
Naive Bayes        & 40.43     (4.10)                           & 39.40         (3.97)        \\
AdaBoost           & 42.94    (3.22)                       & 42.41      (3.94)           \\
MLP                & 50.35   (3.81)                               & 50.67    (4.51)             \\
\hline 
\textbf{TL-MLP }            & 50.76 (4.31)                            & \textbf{53.60} (4.43)             \\
\textbf{TL-MLP-C*}   &   \textbf{54.50} (4.16)          & \textbf{55.12} (4.14)               \\
\bottomrule
\end{tabular}
\end{table}

Table~\ref{tab:result} shows the performance of different thermal comfort modelling algorithms. We use all the ten features described in Section~\ref{subsec:featureselection} on most algorithms except for the PMV model. From Table~\ref{tab:result}, we can see that the PMV model only performs better than the Random baseline and SVM classifiers (kernel = 'Linear') in accuracy. The F1-score of linear SVM is still higher than the PMV model. This may be due to the fact we use more features in machine learning classifiers while the PMV model only has six factors. We will discuss the prediction performance with different features sets later in Section~\ref{sec:improve}. 

From Table~\ref{tab:result}, it can be observed that the Random Forest algorithm performs best on all metrics compared with the PMV model, random baseline and other data-driven models including eight traditional machine learning classifiers. This may because Random Forest is usually regarded as the best classification algorithm for small datasets \cite{hu2019heterogeneous} and has been proved to have the highest prediction accuracy for thermal sensation \cite{luo2020comparing}.


Most importantly, we find that TL-MLP has higher F1-score for thermal comfort classification than other machine learning methods without using transfer learning. Though TL-MLP has better prediction performance than MLP on all metrics, the prediction accuracy of TL-MLP is slightly lower than the Random Forest. The potential reason is that TL-MLP transfer knowledge from all HVAC buildings in the world regardless of the different climate zones, leading to low prediction accuracy than Random Forest. Excitingly, transfer learning based thermal comfort model from the same climate zone (TL-MLP-C*) works better than all state-of-the-art algorithms on all metrics (accuracy and F1-score), indicating the effectiveness of the proposed approach. 

\begin{figure}
  \centering
  \subfigure[MLP\label{fig:cm1}]{\includegraphics[width=0.45\linewidth]{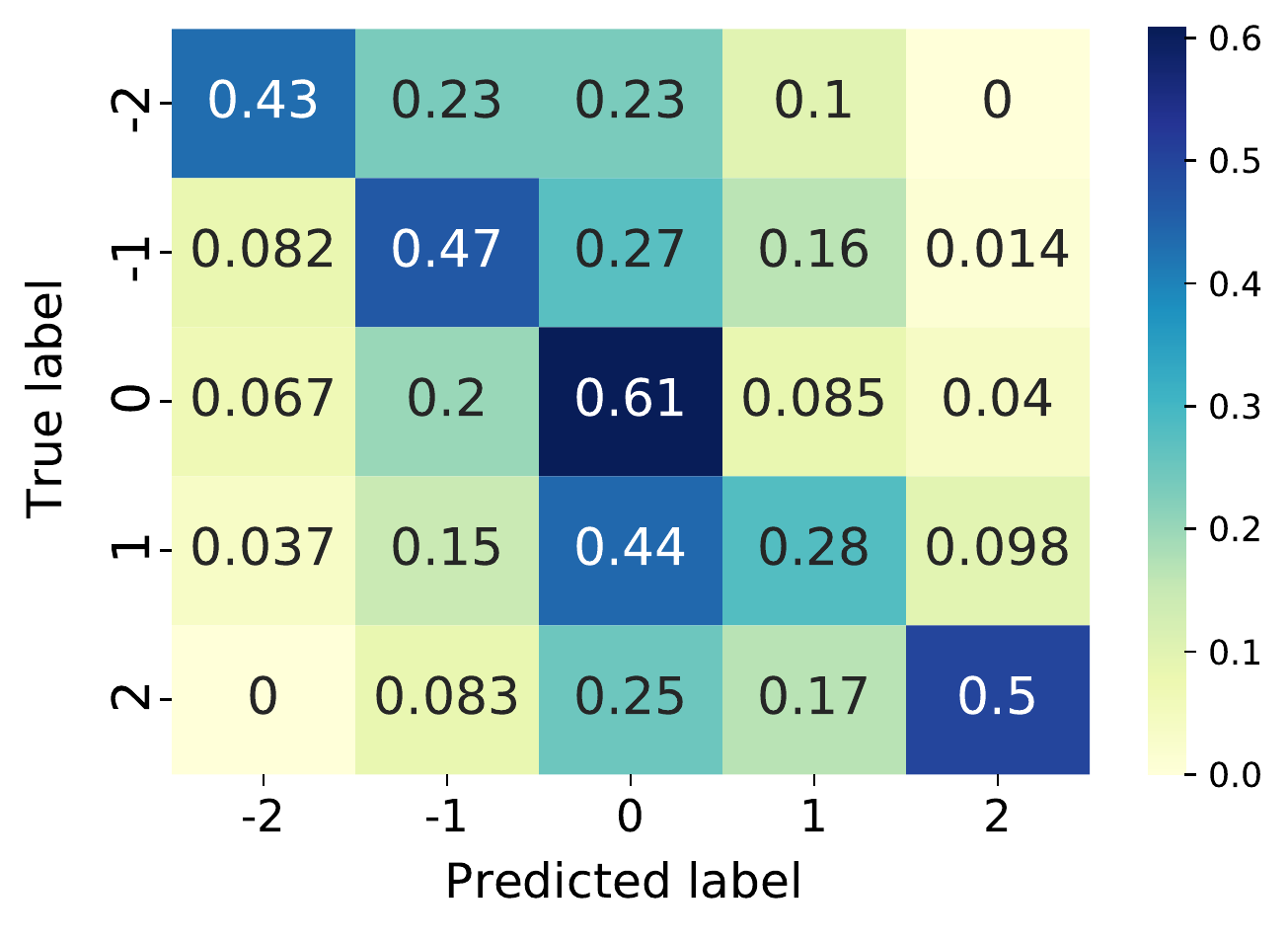}}
  \hspace{0.2cm}
  \subfigure[TL-MLP-C* (Same Climate Zone)\label{fig:cm2}]{\includegraphics[width=0.45\linewidth]{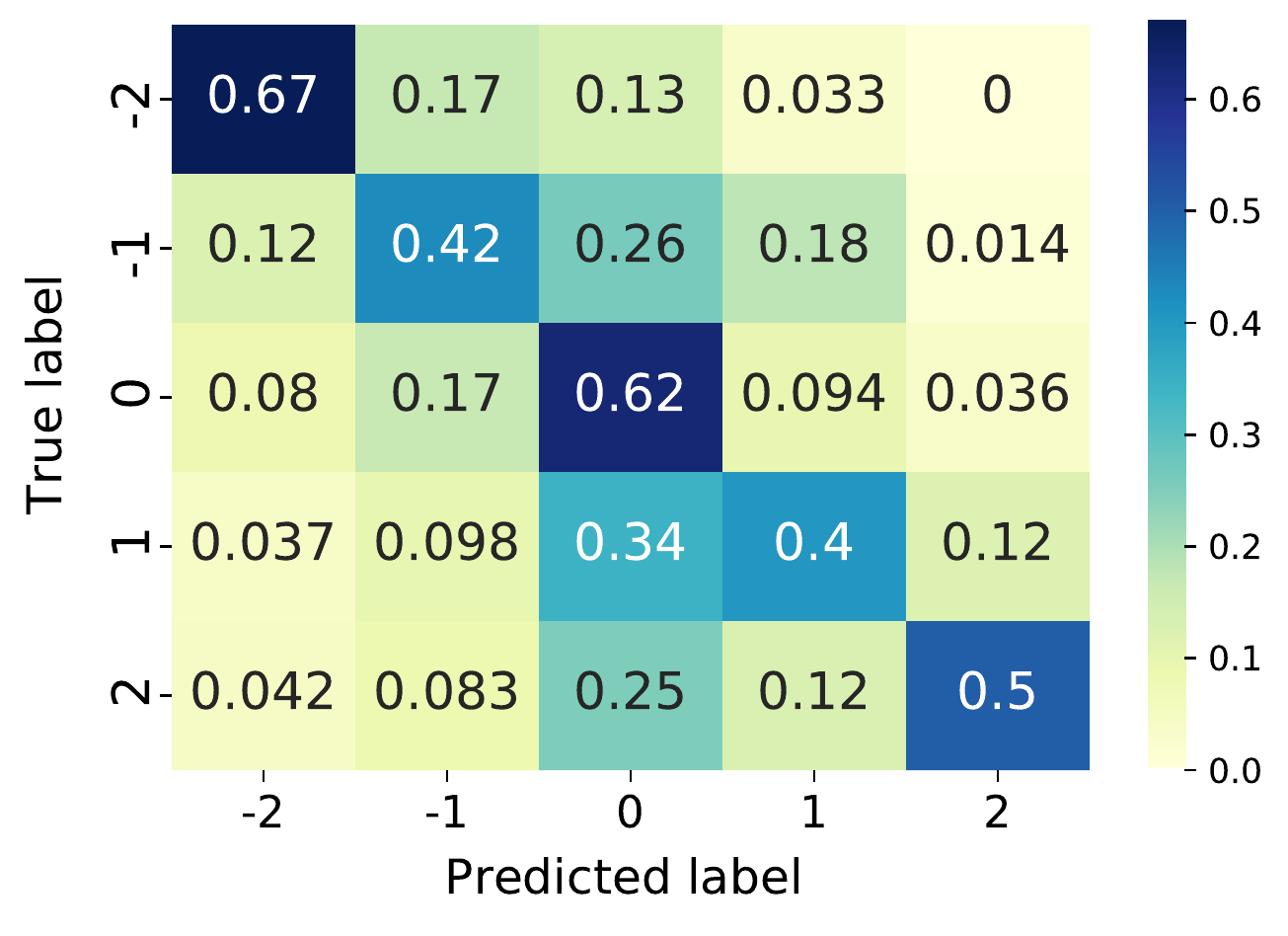}}
  \caption{confusion matrix on the target domain}
  \label{fig:cm}
\end{figure}

To further investigate how the proposed TL-MLP-C* improves the prediction performance in comparison to MLP, we show the confusion matrix diagrams for MLP and TL-MLP-C* in Figure~\ref{fig:cm}. It can be observed that the MLP model can predict the label 0 (neutral) with the highest probability 0.61 which is similar to 0.62 in TL-MLP-C*. However, it still has high chances to declassify label 1 (slightly warm) to 0 (neutral). Instead, transfer learning based thermal comfort model TL-MLP-C* can predict labels more accurately than traditional MLP model, especially for the minority classes (-2, -1, 1). It can predict 67\% label -2 (cool or cold) and 40\% label 1 (slightly warm) correctly and achieves the average 54.50\% accuracies for all the classes from -2 to 2. 

In summary, our proposed transfer learning based models (TL-MLP and TL-MLP-C*) achieve remarkable performance for thermal comfort prediction compared with the random baseline, traditional PMV model and data-driven algorithms without transfer learning. In particular, the TL-MLP-C* model outperforms state-of-the-art algorithms on both metrics (accuracy and F1-score). At the same time, the improved prediction performance of TL-MLP-C* is significant compared to the standard MLP model. 

\subsection{Impact of Different Feature Combinations}\label{sec:improve}
We will now explore how accurately the proposed TL-MLP and TL-MLP-C* models work when only a set of features is available. Usually, indoor sensors are cheap, unobtrusive and have been installed in many HVAC buildings. However, some features may be unavailable due to factors such as privacy, costs, etc. For instance, occupants may not willing to report their age, which reflects their metabolism level and influence their thermal comfort feelings. Besides, it is somewhat inconvenient to install outdoor weather station outside the building which captures outdoor environment changes (e.g., outdoor air temperature and humidity) more accurately than the official weather stations used for local weather forecasting. 

Hence, in the experiment, we will divide our features into 3 different sets $\mathcal{X}_a,\mathcal{X}_b, \mathcal{X}_c$ based on PMV factors, personal factors and outdoor environmental factors, and then compare the different sets and explore which features contribute most for effective thermal comfort transfer learning. The feature sets are as follows:
\begin{itemize}
    \item $\mathcal{X}_a$: Six basic factors introduced in the PMV model: indoor air temperature, indoor air velocity, indoor relative humidity, indoor radiant temperature, clothing insulation and metabolic rate. This is the most common feature set for thermal comfort modelling used in previous studies \cite{hu2019heterogeneous}.
    \item $\mathcal{X}_b$: Six factors from $\mathcal{X}_a$ and two personal factors: age and gender. Personal factors such as gender and age can be easily collected through background surveys. 
    \item $\mathcal{X}_c$: Eight factors from $\mathcal{X}_b$ and two outdoor environmental factors including outdoor air temperature and outdoor relative humidity. The above two outdoor environmental features need to be accessed from the outdoor weather station near the target building.
\end{itemize}

\begin{table}
\caption{Prediction Performance for Different Feature Sets on the Target Dataset}
\label{tab:featureset}
\begin{tabular}{@{}lllll@{}}
\toprule
\textbf{Sets} & \textbf{Algorithm} & \textbf{Accuracy(\%)}  & \textbf{F1-score(\%)} \\ \midrule
\multirow{3}{*}{$\mathcal{X}_a$}   & PMV                & 33.35                                  & 32.45                 \\
                      & Random Forest & \textbf{34.77}                               & 34.92 \\& MLP              & 33.18 &34.06                  \\
                      & TL-MLP            & 33.53                                 & 35.90                 \\
                      & \textbf{TL-MLP-C*}             & 33.98                              & \textbf{39.32}                 \\
                      [0.1em]\hline \\[-0.7em]
\multirow{2}{*}{$\mathcal{X}_b$}   & Random Forest & 43.43 &43.18\\& MLP                & 42.96                 & 45.31                                  \\
                      & TL-MLP             & 44.10                              & 45.88                \\
                       & \textbf{TL-MLP-C*}             & \textbf{47.10}                              & \textbf{51.15}                 \\
                      [0.1em]\hline \\[-0.7em]
\multirow{2}{*}{$\mathcal{X}_c$}   & Random Forest & 51.41 &52.93\\& MLP                & 50.35                            & 50.67                \\ 
                      & TL-MLP             & 50.76                          & 53.60                 \\ 
                      &\textbf{TL-MLP-C*}             & \textbf{54.50}                            & \textbf{55.12}                 \\
                      \bottomrule 
\end{tabular}
\end{table}

For different feature sets, we use the same oversampling methods and fixed random seeds in neural network training. Table~\ref{tab:featureset} shows the prediction performance for different feature sets on the target dataset. Random Forest and MLP algorithm are chosen to compare with TL-MLP and TL-MLP-C* algorithms due to their relatively high performance showed in Table~\ref{tab:result}. For $\mathcal{X}_a, \mathcal{X}_b, \mathcal{X}_c$ feature sets, we can observe that the performance of TL-MLP and TL-MLP-C* models increase with the growing number of features. In the meantime, TL-MLP-C* model has the highest accuracy and F1-score in each feature set.

For the $\mathcal{X}_a$ feature set, the PMV model works slightly better than the MLP model in accuracy but worse in F1-score metrics. Random Forest algorithm achieves the best performance in accuracy while TL-MLP-C* has the highest F1-score. With transfer learning from source datasets, TL-MLP and TL-MLP-C* have similar prediction accuracy with traditional PMV model. It shows that the advantages of the proposed TL-MLP and TL-MLP-C* models can not be fully utilized when the number of features is limited. 

In the $\mathcal{X}_b$ feature set, all data-driven models achieve better prediction performance than only using the $\mathcal{X}_a$ feature set. This shows that personal information (age and gender) could improve thermal comfort prediction effectively. Moreover, TL-MLP-C* model has the best prediction performance than the other methods in both metrics when considering personal factors.

In comparison to the $\mathcal{X}_a$ and $\mathcal{X}_b$ feature sets, Random Forest, MLP, TL-MLP and TL-MLP-C* work best among all metrics in $\mathcal{X}_c$ feature set. This proves that outdoor environmental changes can affect occupants' thermal sensation in HVAC buildings, and shows the necessity to take outdoor features into consideration for effective thermal comfort modelling.

\subsection{Impact of the Number of Hidden Layers}

\begin{figure}
    \centering
    \includegraphics[width=0.28\textwidth]{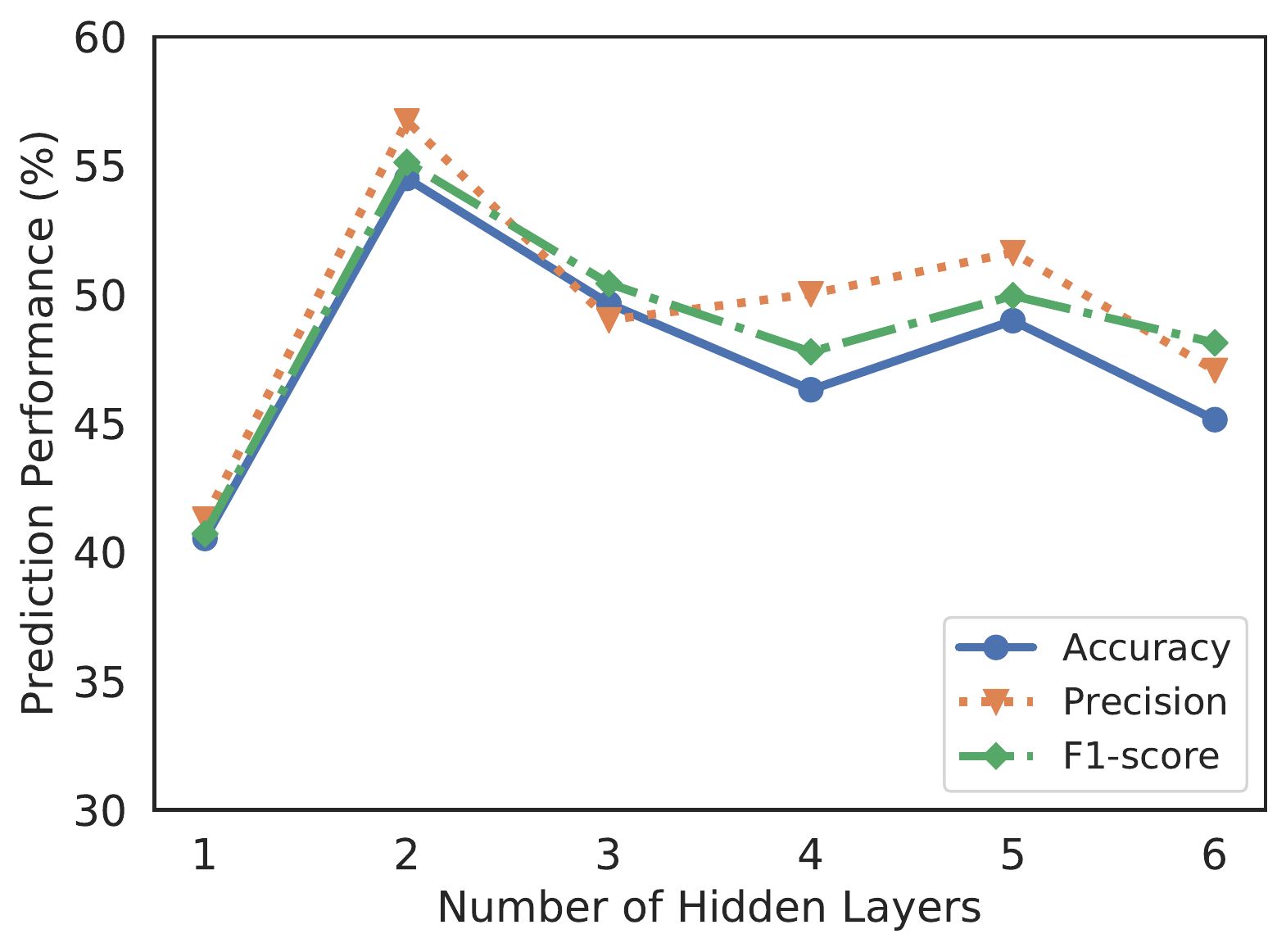}
    \caption{Prediction performance with different number of hidden layers}
    \label{fig:hidden}
\end{figure}

We also conduct adaption experiments by using the different number of hidden layers in TL-MLP-C* model. Figure \ref{fig:hidden} shows the prediction accuracy and F1-score for TL-MLP-C* with the different number of hidden layers. We can observe that the prediction performance is worst in all metrics with only one hidden layer. Since our proposed transfer learning method is to transfer the last layer of the hidden layer, if we only set one hidden layer, the target dataset will have little contribution to the prediction model. When hidden layers are set to 2, the proposed TL-MLP-C* model has the highest prediction performance in accuracy and F1-score. As the number of hidden layers continues to increase, the prediction performance tends to decrease, which may due to the model being overfitting with more trainable parameters.

Finally, despite that our proposed TL-MLP-C* model has better thermal comfort prediction performance than the-state-of-art methods, the achieved accuracy (54.50\%) is still not remarkably high. There are several potential reasons: (1) We adopt \textit{TabularGAN} to re-sample the minority classes for meaningful classification. 50\% size of instances in each class were synthesized while ensuring that the number of samples per category does not exceed the number of samples in the majority class. Though some previous works achieve slightly higher accuracy for thermal comfort prediction (e.g., 63.09\% in \cite{hu2019heterogeneous}, 62\% in \cite{luo2020comparing}), they only assigned a bit higher weights to the instances in the minority classes which can not handle the class imbalance problem as well as we do. (2) Predicting thermal comfort is challenging since there are many factors that affect occupant thermal sensation (as discussed in Section \ref{sec:intro}). There may also exist lots of response bias during the survey. Therefore, the classification accuracy in most previous research is not good too and rarely higher than 60\% even for personal thermal comfort modelling; (3) It could be better to regard the thermal comfort prediction as a regression problem instead of the classification problem. For example, classifying '-2' (cool) to '-1' (slightly cool) should be more acceptable than classifying '-2' (cool) to '+2' (warm). We will try thermal comfort regression in future work.
\section{Conclusion}
\label{sec:conclusion}

A huge amount of sensor data has been generated in cities all over the world. Utilising the sensor data from multiple cities to benefit the target city has become a critical issue in recent years. In this research, we have answered the following research question: \textit{can we predict occupants' thermal comfort accurately by learning from multiple buildings in the same climate zone when we do not have enough data? If so, which features contribute most for effective thermal comfort transfer learning?} We proposed the transfer learning based thermal comfort modelling and applied a generative adversarial network based resampling method for meaningful thermal comfort classification. 

By retaining the last hidden layer of the neural network from the source domain (ASHARE RP-884 dataset and the Scales Project dataset), we trained the thermal comfort model for the Friends Center building from the Medium US Office dataset and found the optimal parameter settings for lower hidden layers. Extensive experimental results showed that the proposed TL-MLP and TL-MLP-C* models outperform the state-of-the-art algorithms for thermal comfort prediction. 
Interestingly,  the most significant feature sets are identified for effective thermal comfort transfer learning. 

This research provides a significant view for learning thermal comfort related sensor data from multiple cities in the same climate zone to benefit thermal comfort prediction in the target building with limited data. In the future, we plan to propose more advanced transfer learning techniques to find the transferable representations between the source domain and target domain. Also, we would like to utilize occupant behaviours (e.g., heating, drinking) to improve the performance of thermal comfort prediction.
    
\section*{Acknowledgement}
This research was supported by the Australian Government through the Australian Research Council's Linkage Projects funding scheme (project LP150100246). We also sincerely thank the anonymous reviewers.

\bibliographystyle{ACM-Reference-Format}
\bibliography{Nan}

\end{document}